\documentclass{article} 
\usepackage{iclr2026_conference,times}

\usepackage{amsmath,amsfonts,bm}

\def\eqref#1{equation~\ref{#1}}

\def\1{\bm{1}}

\DeclareMathAlphabet{\mathsfit}{\encodingdefault}{\sfdefault}{m}{sl}
\SetMathAlphabet{\mathsfit}{bold}{\encodingdefault}{\sfdefault}{bx}{n}

\usepackage{hyperref}
\usepackage{url}
\usepackage{xspace}
\usepackage{booktabs}   
\usepackage{adjustbox}  
\usepackage{makecell}   
\usepackage[table,xcdraw]{xcolor}
\usepackage[dvipsnames]{xcolor}
\definecolor{lightgrey}{HTML}{E7E7E7}
\usepackage{multirow} 
\usepackage{amsmath,amssymb}      
\usepackage{algorithm}            
\usepackage[noend]{algpseudocode} 
\usepackage{float}                
\usepackage{array}
\usepackage{tabularx}
\usepackage[position=bottom]{caption}
\usepackage{longtable}
\usepackage{ragged2e} 

\newcommand{\ci}[3]{#1_{#2}^{#3}}

\title{\ours: Reasoning with Episodic Memory \\in Language Agents}

\usepackage{etoolbox}

\makeatletter
\patchcmd{\@maketitle}{\rule{\z@}{24pt}}{\rule{\z@}{8pt}}{}{}
\patchcmd{\@maketitle}{\rule{\z@}{24pt}}{\rule{\z@}{8pt}}{}{}
\patchcmd{\@maketitle}{\rule{\z@}{24pt}}{\rule{\z@}{8pt}}{}{}
\makeatother

\iclrfinalcopy

\author{Yiheng Shu$^{1}$, Saisri Padmaja Jonnalagedda$^{2}$, Xiang Gao$^{2}$, Bernal Jiménez Gutiérrez$^{1}$ \AND
Weijian Qi$^{1}$, Kamalika Das$^{2}$, Huan Sun$^{1}$, Yu Su$^{1}$\\[2pt]
{\normalfont $^{1}$The Ohio State University \quad $^{2}$Intuit AI Research}\\
{\normalfont\texttt{\{shu.251, su.809\}@osu.edu, \{saisri\_jonnalagedda, xiang\_gao\}@intuit.com}}
}

\newcommand{\ours}{REMem\xspace}

\usepackage{etoc}
\usepackage{titletoc}
\usepackage{svg}
\newcommand\AppendixTOC{ \startcontents \printcontents{}{1}{\noindent\textbf{\Large Appendix Contents}\vspace{4pt}} }

\begin{document}

\maketitle

\begin{abstract}

Humans excel at remembering concrete experiences along spatiotemporal contexts and performing reasoning across those events, i.e., the capacity for episodic memory.
In contrast, memory in language agents remains mainly semantic, and current agents are not yet capable of effectively recollecting and reasoning over interaction histories.
We identify and formalize the core challenges of \textit{episodic recollection} and \textit{reasoning} from this gap, and observe that existing work often overlooks episodicity, lacks explicit event modeling, or overemphasizes simple retrieval rather than complex reasoning. 
We present \ours, a two-phase framework for constructing and reasoning with episodic memory:
1) \textit{indexing}, where \ours converts experiences into a hybrid memory graph that flexibly links time-aware gists and facts.
2) \textit{agentic inference}, where \ours employs an agentic retriever with carefully curated tools for iterative retrieval over the memory graph.
Comprehensive evaluation across four episodic memory benchmarks shows that
\ours substantially outperforms state-of-the-art memory systems such as Mem0 and HippoRAG 2, showing $3.4\%$ and $13.4\%$ absolute improvements on episodic recollection and reasoning tasks, respectively.
Moreover, \ours also demonstrates more robust refusal behavior for unanswerable questions.\footnote{Code and data are available at \url{https://github.com/intuit-ai-research/REMem}.}

\end{abstract}

\section{Introduction}


The ability to intentionally and precisely re-experience events from our past is one of the defining features of human intelligence.
This ``\textit{mental time travel}'' capacity, or \textit{episodic memory}, allows us to access specific events along a spatiotemporal axis, specifying their order, duration, and even causality. 
Unlike \textit{semantic memory}, which stores concepts and knowledge about the world, episodic memory shapes each individual's unique experiences and preferences, establishing a cornerstone for continual learning within one's environment.

Despite its importance for human cognition and rising interest in memory systems for language agents, achieving human-level episodic memory remains elusive, mainly due to the dominance of semantic memory paradigms in current approaches. Parametric memory, embedded in model weights during pre-training or fine-tuning, lacks adaptability and contextual grounding in specific experiences. Model editing methods \citep{editing-llms} can update stored facts but remain limited to modifying static semantic knowledge. 
Retrieval-augmented generation (RAG) systems with embedding models \citep{qwen3embedding,nvembed} enable dynamic knowledge access but still operate in a de-contextualized manner, divorced from spatiotemporal context. 
Finally, more advanced non-parametric systems, which use large language models (LLMs) to construct summaries or semantic graphs \citep{hipporag1, hipporag2, graphrag}, improve over RAG but still prioritize structured world knowledge over lived, interaction-specific experience.

Recently, a few studies have targeted episodic settings more directly. These typically represent episodic memory using entity relationships in (temporal) knowledge graphs with a loss of coherent event contexts \citep{zep} or by selectively inferring what seems important and offering it as a summary \citep{mem0,rmm}, which lacks \textit{explicit event modeling}. 
Critically, they fail to integrate situational dimensions, such as time, location, and participants, within interaction histories. 
Furthermore, when memory must support reasoning across multiple linked events, existing methods rely heavily on similarity-based retrieval, offering limited capacity to infer complex inter-event relationships. 
Given these observations, we believe that a comprehensive event representation, combined with flexible retrieval and reasoning, deserves further exploration.

\begin{figure}[t]
    \centering
    \includegraphics[width=\textwidth]{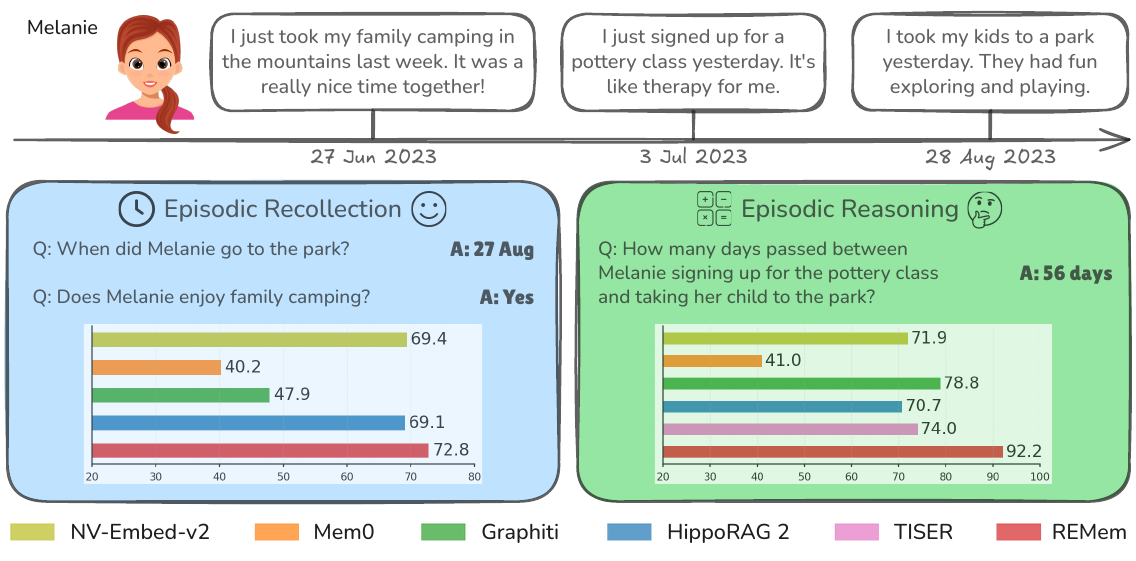}
    \caption{Overview of evaluation on episodic memory. Utterances are grounded to a timeline (top). We evaluate two progressive capabilities and show average scores on each (bottom):
    1) Episodic recollection: recollect temporal and other situational elements of past experiences, measured by LLM-as-a-judge scores on LoCoMo and REALTALK.
    2) Episodic reasoning: multi-step reasoning across the timeline based on recollection, e.g., event-to-event relations, counting, and ordinal queries, measured by LLM-as-a-judge score on Complex-TR and the EM score on Test of Time.
    }
    \label{fig:tasks}
    \vspace{-20pt}
\end{figure}

To advance episodic memory and reasoning for language agents, we first identify two key, progressive challenges, as shown in Figure~\ref{fig:tasks}:
1) \textbf{Episodic recollection}: reconstruct events and their situational dimensions based on experiences, such as time, location, participant, emotion; i.e., the ability to bind situational elements to specific events.
2) \textbf{Episodic reasoning}: multi-step reasoning based on episodic recollection, such as inter-event relations, ordinal constraints, and superlatives.

Then, we introduce a new episodic memory framework, \ours (\underline{R}easoning with \underline{E}pisodic \underline{Mem}ory), for language agents.
We formalize episodic memory as time-aware event representations and propose a \textit{hybrid memory graph} that stores \textbf{gists} (concise, human-readable event summaries with parsed timestamps) and \textbf{facts} (time-scoped triples). 
Unlike existing work that selectively extracts useful information, we clearly \textit{instruct} the LLM to construct memories organized primarily along time and linked to situational dimensions such as participants, locations, and emotions.
We develop an \textit{agentic inference} procedure with carefully curated tools for retrieval and graph exploration. As a result, memory management goes beyond simply matching isolated text spans and enables complex logical composition, including time-range filtering, neighbor exploration, and ordinal constraints.

We conduct one of the most comprehensive evaluations on episodic memory to date, covering four benchmarks on conversational and temporal reading comprehension. 
\ours delivers consistent gains over the current state of the art, achieving a $3.4\%$ and $13.4\%$ absolute improvement on episodic recollection and reasoning tasks, respectively.
\ours also shows unparalleled reasoning capabilities, being the only method to exceed $90\%$ exact match score on the Test of Time benchmark \citep{tot}, and behaves more robustly to refuse unanswerable questions.
These strong results position \ours as a solid step towards bringing effective episodic memory to language agents.

\section{Related Work}
\subsection{Non-Parametric Memory for LLMs}

A broad line of work augments LLMs with non-parametric memory. 
ChatGPT \citep{chatgpt_memory} combines prior chats with user-controlled saved memories for personalization.
Strong embedding models \citep{nvembed,qwen3embedding} yield competitive retrieval, but their flat vector spaces do not explicitly encode episodic or temporal structure. 
Structure-augmented approaches build graphs or memory layers to improve multi-hop and cross-session retrieval \citep{graphrag}: 
HippoRAG 1\&2 \citep{hipporag1,hipporag2} organizes knowledge for associative retrieval and continual updates. 
Graphiti/Zep \citep{zep} maintains temporal knowledge graphs and context-assembly pipelines. 
Mem0 \citep{mem0} extracts and consolidates conversational memory in a graph structure with a knowledge update mechanism.
Orthogonal to graph layering, MemGPT \citep{memgpt} implements OS-style hierarchical paging and virtual context management. 
A-Mem \citep{amem} performs agentic, Zettelkasten-like dynamic linking among notes. 
MemoryBank \citep{memorybank} introduces time-decay–based consolidation and forgetting.
Reflective Memory Management \citep{rmm} further refines what to store and retrieve over time. 
Overall, most existing systems rely on LLM-based summarization or graph construction to extract key information, yet they overlook the crucial role of spatiotemporal context and situational elements in supporting episodic memory. 
These systems rarely propose an explicit design for episodic memory and reasoning.
In contrast, our approach explicitly represents event gists and temporal facts grounded to a timeline, links them to core situational dimensions, and integrates them with tool-augmented reasoning.

\begin{figure}[tb]
    \centering
    \includegraphics[width=\linewidth]{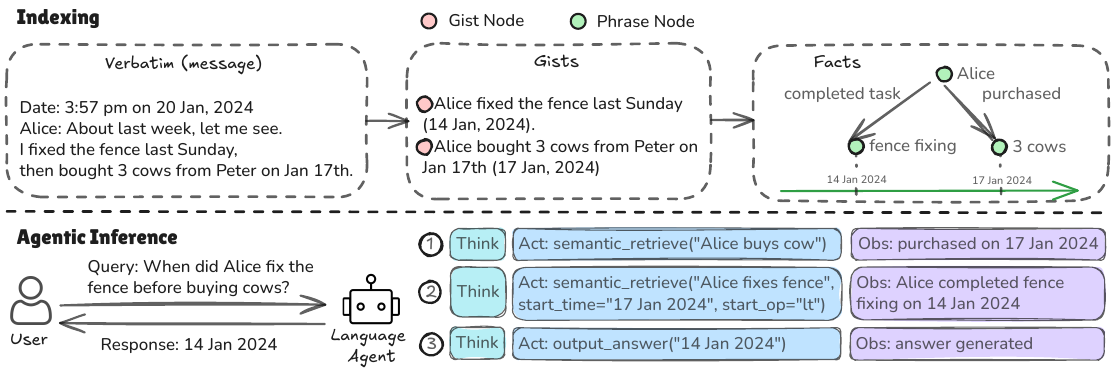}
    \caption{Overview of \ours. The indexing phase turns utterances into time-aware memory by extracting event gists and time-scoped facts (triples) and organizing them as a hybrid graph. The agentic inference phase invokes carefully curated tools over this graph to surface the most relevant gists and facts for reasoning in an iterative manner.}
    \label{fig:method}
\end{figure}

\subsection{Episodic Memory and Reasoning}

Recent benchmarks increasingly stress long-horizon interaction for language agents: LoCoMo \citep{locomo} and REALTALK \citep{realtalk} evaluate multi-session conversational memory in synthetic and real settings.
Conversation understanding of LongMemEval \citep{longmemeval} requires information extraction or multi-session reasoning, but most tasks can be seen as single-retrieval problems only requiring agents to retrieve the correct segments \citep{memoryagentbench}.
Temporal reading comprehension benchmarks, such as TimeQA \citep{timeqa}, MenatQA \citep{menatqa}, Complex-TR \citep{complextr}, and Test of Time \citep{tot}, decompose temporal reasoning into diverse skills: event ordering, date resolution, counting, duration estimation, and timeline construction. 
Methodologically, timeline self-reflection method like TISER \citep{tiser} improves temporal reasoning only through better prompting, while tool-use agents \citep{tremu,tempagent} enable procedural interaction with non-parametric memory. 
Our work complements these efforts by 1) encoding episodes as time-aware gists and facts in a hybrid graph, and 2) using tool-augmented retrieval and graph exploration to perform episodic recollection and reasoning during inference.

\section{Methodology}


Our framework, \ours, allows language agents to reason with episodic memory through a two-phase process: \textit{indexing} and \textit{agentic inference}.
The indexing phase builds a memory graph that stores the gists and facts for episodes.
Then, the agentic inference phase flexibly leverages carefully curated tools to reason over user queries and retrieve from the memory graph.

\begin{table}[tb]
\small
\centering
\caption{Curated tools and their signature. Both the retrieval and graph exploration tools output two sets of results: a list of gists and a list of facts. See Appendix \ref{sec:prompts} for prompts and demonstrations.}
\vspace{0.2em}
\label{tab:tools}
\begin{tabular}{l|l|p{8cm}}
\toprule
\textbf{Type} & \textbf{Name} & \textbf{Arguments} \\ \midrule
\multirow{2}{*}{Retrieval} 
 & semantic\_retrieve & query, start\_time, end\_time, start\_operator, end\_operator \\
 & lexical\_retrieve  & query, start\_time, end\_time, start\_operator, end\_operator   \\ \midrule
\multirow{2}{*}{\makecell[l]{Graph \\ Exploration}} 
 & find\_gist\_contexts  & gist\_id, start\_time, end\_time, start\_operator, end\_operator \\
 & find\_entity\_contexts & subject, object, predicate, start\_time, end\_time, start\_operator, end\_operator, limit, ordering, offset, aggregation \\ \midrule
Flow Control & output\_answer & answer \\ \bottomrule
\end{tabular}
\end{table}

\subsection{Indexing}
\label{subsection:indexing}

Cognitive science reminds us that humans rely more on gist than verbatim memory when making decisions \citep{fuzzytrace}, and our discourse memory is shaped around situation models, which integrate temporal, spatial, causal, and situational dimensions beyond surface details \citep{johnson1983mental,zwaan1998situation}. 
Besides, recent RAG studies show that hybrid memory structures combining concept-level and context-level information are more effective than either alone \citep{graphrag,hipporag2}. 
Motivated by these insights, our principles for constructing a memory graph (Figure \ref{fig:method}) focus on comprehensiveness and flexibility, which jointly encode gists with multiple situational dimensions and facts with temporal contexts.
Specifically, we perform the following steps.

\textbf{1) Gist Extraction}:
For each event statement or chat session, we generate one or multiple gist statements in natural language. 
Each gist is prefixed with the episode's timestamp (reference time) if applicable, and any relative temporal expressions are resolved to absolute dates.
For each episode (e.g., a chat session), this yields a gist list, where each gist is a concise sentence capturing the episode’s key details, including participants, actions, objects, locations, intentions, and quantities, in a single atomic event description.
\textbf{2) Fact Extraction}:
We further extract structured facts from each episode's text and the extracted gist list.
These facts are represented as (subject, predicate, object) triples, where each field is a schemaless phrase, mainly capturing who did what to whom.
We also extract temporal contexts (dates/times) and ground each fact in a timeline by optionally attaching Wikidata-style qualifiers \citep{wikidata}, \textit{point\_in\_time}, \textit{start\_time}, and \textit{end\_time}, to the corresponding triple.
We preserve gists and facts as they are added over time, even when they are potentially contradictory, thereby maintaining a record that can be revisited historically.
\textbf{3) Graph Construction}:
Using the above outputs, we construct a memory graph that integrates gist nodes and phrase nodes.
Gist nodes serve as \textit{context-level} episodic representations, each connected to the phrase nodes extracted from the same chunk.
At the \textit{concept-level}, subject and object phrase nodes of each fact are directly linked by edges, encoding the extracted relationships between those phrases.
Thus, a \textit{hybrid} memory graph combining concept and context levels is formed.
To further enhance connectivity, we add synonymy edges between gist nodes whose embedding similarity exceeds a threshold, following the HippoRAG 2 approach \citep{hipporag2}.
This mechanism clusters semantically related gists (e.g., different phrasings of similar events) and enriches the graph with higher-level semantic connections.

\subsection{Agentic Inference}
\label{subsection:agentic_inference}

In contrast to the common RAG approach of simply performing a one-time matching with related text, our inference phase includes \textit{iterative retrieval} to handle complex logic, and \textit{mental time travel}, which filters memory entries according to temporal conditions.
We adopt a ReAct-style agent \citep{react,middleware} equipped with three categories of carefully curated tools to access the hybrid graph: 1) retrieval tools, 2) graph exploration tools, and 3) the flow control tool.
The results returned by retrieval or graph exploration tools contain both gists and facts to provide a comprehensive view.
Specifically, the agent follows a three-stage protocol over the hybrid memory graph, as shown in Table \ref{tab:tools}. 
Full tool specifications are provided in Appendix~\ref{sec:prompts}.

\textbf{1) Retrieval}:
The agent first invokes \textit{semantic\_retrieve} (using an embedding model) or \textit{lexical\_retrieve} (using BM25) tool to obtain truncated seed nodes and their contexts from the graph, e.g., candidate entity IDs, temporal windows, and coarse topical cues. The agent decomposes complex questions into simpler sub-queries that guide the next step.
\textbf{2) Graph Exploration}:
Using the seed nodes retrieved from stage 1, the agent issues targeted calls to \textit{find\_gist\_contexts} to obtain episode-level narratives and temporally grounded evidence, or uses \textit{find\_entity\_contexts} when the query explicitly targets entities under a known graph schema.
\textit{find\_entity\_contexts} not only supports specifying subjects, predicates, or objects, but also filters gists or facts that meet specific temporal conditions.
\textbf{3) Flow Control}:
Once sufficient evidence has been gathered, the agent invokes \textit{output\_answer} to generate the final response upon reaching confidence. This procedure exploits the entire interaction history, incorporating both explored gists and facts, to conduct the concluding reasoning.

\section{Experimental Setup}

\begin{table}[tb]
\small
\centering
\caption{The statistics of sampled datasets.}
\vspace{0.2em}
\begin{tabular}{l|rrrrrr}
\toprule
Datasets & LoCoMo & REALTALK & Complex-TR & Test of Time\\ \midrule
\# of queries & $1,986$ & $728$ & $1,000$ & $2,800$ \\
\# of chunks/messages & $5,882$ & $8,944$ & $1,095$ & $124,919$  \\
\# of graphs & $10$ & $10$ & $1$ & $560$  \\
\bottomrule
\end{tabular}
\label{tab:dataset_stat}
\end{table}

\subsection{Datasets}
\label{subsec:datasets}

For the episodic recollection task, we use conversational question answering (QA) benchmarks: \textbf{LoCoMo} \citep{locomo} is a synthesized conversational benchmark, while \textbf{REALTALK} \citep{realtalk} is collected from real human conversations.
Both benchmarks include questions that cover temporal aspects and non-temporal situational aspects, and we use all samples from them.
For the episodic reasoning task, we use reading comprehension benchmarks \textbf{Complex-TR} \citep{complextr} and \textbf{Test of Time} \citep{tot}. 
From Complex-TR, we randomly sample $1,000$ queries.
From the Test of Time, we adopt all samples from its semantic part, since this part explicitly requires memory.

\subsection{Methods for Comparison}

We select state-of-the-art methods from the following categories for comparison:
1) Strong embedding models from the MTEB benchmark \citep{mteb} for RAG setting, including \textbf{Qwen/Qwen3-Embedding-8B} \citep{qwen3embedding} and \textbf{nvidia/NV-Embed-v2} \citep{nvembed}.
2) Structure-augmented memory approaches, including \textbf{Mem0} \citep{mem0}, \textbf{Graphiti} \citep{zep}, and \textbf{HippoRAG 2} \citep{hipporag2}, where the first two methods are originally evaluated on episodic memory tasks.
We reproduced the experiments using their \textit{open-source} implementation rather than proprietary ones.
3) The prompt method \textbf{TISER} for temporal reasoning  \citep{tiser}: give the query and retrieved contexts, we use its prompt for the final generation. 
The method is orthogonal to any memory system.
4) Additional references: \textbf{Oracle Message} takes the oracle retrieval results and only performs generation.
\textbf{Full-Context} uses the entire corpus and the query for generation.
Due to its limited context window, it should not be regarded as a memory method, but we take it as a reference rather than a comparative memory method.

\subsection{Metrics}
\label{subsec:metrics}

For Test of Time \citep{tot}, we use the unique metric exact match (\textbf{EM}) score.
For the remaining benchmarks, we use token-based \textbf{F1}, \textbf{BLEU-1} \citep{bleu}, and \textbf{LLM-as-a-judge} scores (denoted as LLM-J later) as metrics for QA tasks.
We adopt the F1 calculation from HippoRAG 2 \citep{hipporag2}, BLEU-1 implementation from HuggingFace Evaluate \citep{hf_evaluate}. 
We adopt the same LLM evaluation prompts as Mem0 \citep{mem0}, which account for both temporal and conversational contexts.

\subsection{Implementation Details}
\label{subsec:implementation-details}

We use GPT-4.1-mini-2025-04-14 \citep{openai_gpt4_1_mini} as the default LLM and nvidia/NV-Embed-v2 \citep{nvembed} as the default embedding model for both extraction and QA tasks, in \ours as well as in comparison methods.
For baselines using the embedding model, we retrieve the top-$10$ original passages (messages in conversation).
For Mem0 and Graphiti, we retrieve the top-$10$ of their processed chunks.
\ours applies the same retrieval scope for each step, operating over the top-$10$ gists and facts.
We use the top-$3$ returned sessions from HippoRAG 2 for final generation.

We explore two settings for our method: \textbf{\ours-I}(terative) autonomously selects tools in a multi-step retrieval and reasoning process, following the protocol in \S \ref{subsection:agentic_inference}. \textbf{\ours-S}(ingle) only adopts a single-step embedding-based retrieval and then generation.
The maximum number of agentic inference steps in \ours-I for each dataset is selected from $2$ to $5$ based on a small validation set. Then, we use $3$ for episodic recollection tasks and $5$ for episodic reasoning tasks.
The similarity threshold for synonymy edges is set to $0.8$, following HippoRAG 2.

\begin{table}[tb]
\centering
\small
\caption{Performance (\%) on episodic recollection task. 
The highest value and second-highest value in each column are bold and underlined, respectively.
Numbers are means with 95\% bootstrap confidence intervals as subscripts and superscripts. The same applies to the tables below.
}
\vspace{0.2em}
\begin{tabular}{l|ccc|ccc}
\toprule
 & \multicolumn{3}{c|}{LoCoMo ($1,986$)} & \multicolumn{3}{c}{REALTALK ($728$)}  \\
 \cmidrule(lr){2-4} \cmidrule(lr){5-7} 
Methods & F1 & BLEU-1 & LLM-J & F1 & BLEU-1 & LLM-J  \\ \midrule
\rowcolor{gray!15}Oracle Message & $48.0$ & $38.3$ & $81.0$ & $-$ & $-$ & $-$ \\
\rowcolor{gray!15}Full-Context   & $37.8$ & $28.6$ & $76.7$ & $25.3$ & $18.6$ & $65.1$\\
\midrule
\rowcolor{gray!8}
\multicolumn{7}{c}{\textit{\textbf{Large Embedding Models}}}\\
Qwen3-Embed-8B   & $\ci{35.3}{-1.6}{+1.7}$ & $\ci{28.9}{-1.5}{+1.8}$ & $\ci{64.2}{-2.0}{+2.4}$ &
                    $\ci{20.2}{-1.6}{+1.8}$ & $\ci{14.9}{-1.4}{+1.6}$ & $\ci{52.5}{-3.6}{+3.4}$  \\
NV-Embed-v2 (7B) & $\ci{39.6}{-1.4}{+1.7}$ & $\ci{31.0}{-1.4}{+1.7}$ & $\ci{73.0}{-1.8}{+2.0}$ &
                    $\ci{23.8}{-1.8}{+1.9}$ & $\ci{17.7}{-1.5}{+1.5}$ & $\ci{59.5}{-3.6}{+3.3}$ \\ 
\rowcolor{gray!8}
\multicolumn{7}{c}{\textit{\textbf{Structure-Augmented Memory}}} \\
Mem0        & $\ci{25.1}{-1.5}{+1.7}$ & $\ci{18.0}{-1.1}{+1.4}$ & $\ci{49.7}{-2.2}{+2.3}$ &
               $\ci{9.8}{-1.3}{+1.5}$  & \ \ $\ci{7.2}{-1.0}{+1.1}$  & $\ci{14.3}{-2.3}{+2.7}$ \\
Graphiti    & $\ci{33.7}{-1.8}{+1.8}$ & $\ci{28.9}{-1.7}{+1.9}$ & $\ci{52.5}{-2.3}{+2.3}$ &
               $\ci{15.1}{-1.5}{+1.8}$ & $\ci{11.5}{-1.2}{+1.4}$ & $\ci{35.3}{-3.3}{+3.7}$ \\
HippoRAG 2  & $\ci{39.0}{-1.6}{+1.6}$ & $\ci{30.8}{-1.5}{+1.5}$ & $\ci{74.0}{-2.1}{+1.7}$ &
               $\ci{21.9}{-1.6}{+1.6}$ & $\ci{16.2}{-1.3}{+1.4}$ & $\ci{55.8}{-3.6}{+3.4}$  \\
\rowcolor{gray!8}
\multicolumn{7}{c}{\textit{\textbf{Ours}}}  \\
\ours-I & $\ci{\mathbf{42.4}}{-1.6}{+1.6}$ & $\ci{\mathbf{32.7}}{-1.6}{+1.5}$ & $\ci{\underline{76.2}}{-1.9}{+2.0}$ &
           $\ci{\underline{25.6}}{-1.7}{+1.8}$ & $\ci{\underline{18.1}}{-1.4}{+1.6}$ & $\ci{\underline{63.7}}{-3.5}{+3.6}$ \\
\ours-S & $\ci{\underline{41.3}}{-1.5}{+1.6}$ & $\ci{\underline{31.5}}{-1.4}{+1.6}$ & $\ci{\mathbf{77.5}}{-1.6}{+1.9}$ &
           $\ci{\mathbf{26.2}}{-1.6}{+1.8}$ & $\ci{\mathbf{19.2}}{-1.3}{+1.5}$ & $\ci{\mathbf{65.3}}{-3.1}{+3.6}$ \\ 
\bottomrule
\end{tabular}
\label{tab:episodic-recollection}
\vspace{-15pt}
\end{table}

\begin{table}[t]
\centering
\small
\caption{Performance (\%) on episodic reasoning tasks.}
\vspace{0.2em}
\label{tab:episodic-reasoning}
\begin{tabular}{l|ccc|c}
\toprule
 &  \multicolumn{3}{c|}{Complex-TR ($1,000$)} & Test of Time ($2,800$) \\
\cmidrule(lr){2-4} \cmidrule(lr){5-5}
Methods &  F1 & BLEU-1 & LLM-J & EM \\ \midrule
\rowcolor{gray!15}
Full-Context  & $74.2$ & $68.0$ & $81.6$ & $79.7$ \\\midrule
\rowcolor{gray!8}
\multicolumn{5}{c}{\textit{\textbf{Large Embedding Models}}} \\
Qwen3-Embed-8B   & $\ci{77.1}{-2.1}{+2.3}$ & $\ci{71.4}{-2.4}{+2.5}$ & $\ci{80.9}{-2.5}{+2.5}$ & $\ci{70.3}{-1.7}{+1.8}$ \\
NV-Embed-v2 (7B) & $\ci{77.5}{-2.1}{+2.2}$ & $\ci{71.9}{-2.3}{+2.3}$ & $\ci{80.4}{-2.5}{+2.6}$ & $\ci{68.9}{-1.7}{+1.7}$ \\
\ \ \ \ w/ TISER & $\ci{\underline{88.1}}{-1.5}{+1.7}$ & $\ci{\underline{83.6}}{-1.8}{+2.2}$ & $\ci{88.3}{-1.9}{+1.9}$ & $\ci{68.9}{-1.8}{+1.8}$ \\
\rowcolor{gray!8}
\multicolumn{5}{c}{\textit{\textbf{Structure-Augmented Memory}}} \\
Mem0       & $\ci{43.1}{-2.7}{+2.8}$ & $\ci{35.1}{-2.4}{+2.5}$ & $\ci{41.0}{-3.0}{+3.0}$ & $-$ \\
Graphiti   & $\ci{76.6}{-2.3}{+2.2}$ & $\ci{71.4}{-2.5}{+2.4}$ & $\ci{78.8}{-2.6}{+2.6}$ & $-$ \\
HippoRAG 2 & $\ci{78.2}{-1.8}{+2.3}$ & $\ci{72.7}{-2.4}{+2.4}$ & $\ci{81.5}{-1.3}{+3.4}$ & $\ci{66.9}{-1.7}{+1.7}$ \\
\rowcolor{gray!8}
\multicolumn{5}{c}{\textit{\textbf{Ours}}} \\
\ours-I          & $\ci{83.3}{-1.8}{+1.8}$ & $\ci{77.6}{-2.1}{+2.2}$ & $\ci{\underline{89.6}}{-2.0}{+2.0}$ &
                    $\ci{\mathbf{93.1}}{-1.1}{+0.9}$  \\
\ \ \ \ w/ TISER & $\ci{\mathbf{90.6}}{-1.4}{+1.2}$ & $\ci{\mathbf{86.0}}{-1.7}{+1.7}$ & $\ci{\mathbf{92.0}}{-1.7}{+1.6}$ &
                    $\ci{\underline{90.6}}{-1.2}{+1.0}$ \\
\ours-S          & $\ci{78.5}{-2.1}{+2.0}$ & $\ci{72.7}{-2.4}{+2.4}$ & $\ci{82.6}{-2.4}{+2.3}$ &
                    $\ci{72.5}{-1.8}{+1.8}$ \\
\bottomrule
\end{tabular}
\vspace{-10pt}
\end{table}

\section{Results}

\subsection{Episodic Recollection}

The results on episodic recollection tasks are shown in Table \ref{tab:episodic-recollection}.
Overall, REALTALK collected from human utterances poses greater challenges than synthetic LoCoMo.
As a reference, Full-Context performs well, but the inference cost is notable: the average number of input tokens for each LoCoMo query is $26$k, while \ours-I and \ours-S consume $9$k and $0.9$k tokens, respectively, during the inference phase.
The RAG method using large embedding models serves as a strong baseline, especially NV-Embed-v2.
Structure-augmented memory methods demonstrate subpar overall performance, particularly on REALTALK, where human utterances are more spontaneous, noisy, and less dense in information.
Mem0 extracts many statements, but most are subsequently discarded from memory by its own decision, resulting in only a few details being remembered.
Graphiti constructs a temporal knowledge graph centered around entities, which results in the loss of coherent contextual information related to various situational dimensions of events.
HippoRAG 2 lacks any modeling of the temporal dimension or events. 
Its overall performance is only comparable to the embedding baseline due to its use of passage nodes with embedding-based retrieval.
In contrast, \ours-I and \ours-S substantially outperform existing methods and even approach the oracle performance.
More detailed results for LoCoMo and REALTALK are provided in Appendix \ref{subsec:results-on-locomo} and \ref{subsec:results-on-realtalk}, respectively, where we show \ours-I and \ours-S have their respective strengths across different metrics. 
Notably, cross-session questions in LoCoMo account for only $14.2\%$ of the benchmark, and support messages for most questions originate from a single chat session, which explains why \ours-S as a single-step variant performs better in certain scenarios.

\subsection{Episodic Reasoning}
\label{subsec:episodic-reasoning}

The results on episodic reasoning tasks are shown in Table \ref{tab:episodic-reasoning}.
RAG using large embedding models remains a strong baseline.
TISER, as a prompt guiding language agents to perform temporal reasoning, demonstrates a stronger reasoning capability on Complex-TR, compared to our straightforward answer-generation prompts (Figure \ref{fig:tool description 1}). 
However, it remains a fixed prompt that struggles to cover all episodic reasoning challenges, and excels at chronological questions, such as before/after or first/last types.
The structure-augmented memory methods mainly leverage embedding to retrieve at various granularities, e.g., entities or summaries. 
But this simple semantic matching is insufficient for capturing the full contextual information necessary for complex reasoning, and it fails to support the required logical operations.
\ours shows absolute superiority in complex reasoning tasks.
In particular, \ours-I demonstrates a clear advantage over \ours-S (LLM-J $+7.0$, EM $+20.6$) and becomes the \textbf{unique one exceeding $90\%$ EM score}, benefiting from multi-step retrieval and flexible tool use.
Moreover, \ours-I obtains substantially larger improvements over Full-Context (LLM-J $+8.0$, EM $+13.4$) on these challenging episodic reasoning tasks, compared to that on episodic recollection tasks.
More detailed results for Complex-TR and Test of Time are provided in Appendix \ref{subsec:results-on-complex-tr} and \ref{subsec:results-on-tot}, respectively.

\begin{table}[tb]
\centering
\small
\caption{Ablation study on LoCoMo and Complex-TR, regarding the graph structure and the usage of retrieval tools.}
\vspace{0.2em}
\addtolength{\tabcolsep}{-3pt}
\begin{tabular}{l|ccc|ccc}
\toprule
 & \multicolumn{3}{c|}{LoCoMo} & \multicolumn{3}{c}{Complex-TR} \\
 \cmidrule(lr){2-4} \cmidrule(lr){5-7} 
Methods & F1 & BLEU-1 & LLM-J &  F1 & BLEU-1 & LLM-J \\ \midrule
\ours-S & $41.3$ & $31.5$ & $\mathbf{77.5}$ & $78.5$ & $72.7$ & $82.6$ \\
\ours-I & $\mathbf{42.4}$ & $32.7$ & ${76.2}$ & $\mathbf{83.3}$ & $\mathbf{77.6}$ & $\mathbf{89.6}$ \\
\ \ \ \ w/o Gists & $31.7$ & $28.7$ & $48.9$ & $80.3$ & ${75.9}$ & $80.9$ \\
\ \ \ \ w/o Facts & ${42.0}$ & ${32.6}$ & $74.1$ & ${80.5}$ & $74.5$ & $87.2$ \\
\ \ \ \ w/o Synonymy edges & $37.6$ & $28.7$ & $76.4$ & $81.6$ & $75.6$ & $89.2$   \\
\ \ \ \ w/o Tool semantic\_retrieve & $41.7$ & $\mathbf{33.3}$ & $72.8$ & $82.4$ & $76.4$ & $88.1$ \\
\ \ \ \ w/o Tool lexical\_retrieve & $40.6$ & $31.2$ & $76.8$ & $81.7$ & $75.8$ & $87.5$ \\ 
\bottomrule
\end{tabular}
\label{tab:ablation}
\vspace{-10pt}
\end{table}

\section{Discussions}

\subsection{Ablation Study}

We conducted ablation experiments on LoCoMo and Complex-TR (Table \ref{tab:ablation}). 
Both gists and facts are important, but they contribute in different ways.
Removing gists leads to the largest degradation, especially on LoCoMo, where LLM-J drops from $76.2$ to $48.9$.
This supports our design choice that gists carry the main situational elements.
Removing facts produces a smaller yet consistent drop, particularly on Complex-TR, where LLM-J decreases from $89.6$ to $87.2$.
This suggests that facts play a supporting role for multi-hop reasoning, by providing concrete anchors that connect concepts across sessions (Appendix~\ref{subsec:appendix-ablation-study}).
Graph structure and retrieval tools also matter.
Ablating synonymy edges reduces F1 and BLEU-1 on both datasets, indicating that modeling synonymic relationships improves lexical robustness and recall, while LLM-J remains almost unchanged, which implies that the core reasoning path is largely preserved.
Finally, removing either the semantic or the lexical retrieval tool hurts performance.
Without semantic\_retrieve, LLM-J on Complex-TR decreases from $89.6$ to $88.1$ and also drops on LoCoMo, showing that semantic retrieval is crucial for finding conceptually relevant memories.
Without lexical\_retrieve, F1 and BLEU-1 decline and LLM-J on Complex-TR falls to $87.5$, indicating that lexical retrieval complements semantic retrieval by improving coverage of surface forms.

\subsection{Refusal Performance}
\label{subsec:refusal-performance}

In real-world applications, users will always pose queries for which the system lacks sufficient context. 
Such adversarial or unanswerable questions are still valid for evaluation purposes, and we include them as part of the full LoCoMo benchmark.
For these cases, LoCoMo labels them as ``\textit{no information available}''. 
Therefore, if a method either fails to produce an answer or explicitly outputs this phrase as instructed, we take it as a refusal to answer.
For these metrics of refusal behaviors (not QA metrics) as shown in Table~\ref{tab:unanswerable-detection}, \ours achieves the highest F1 ($63.96\%$) by coupling the best precision ($73.3\%$) with competitive recall ($56.8\%$).
Compared to Graphiti, which produces $954$ refusals with low precision ($38.9\%$) despite the highest recall ($83.6\%$), \ours improves precision by $+34.4$ points and F1 by $+10.9$ points, while producing roughly one-third as many refusals ($344$ vs.\ $954$), indicating substantially fewer unnecessary rejections. 
Compared to Mem0, which is overly permissive (only $90$ refusals), \ours correctly flags more unanswerable cases. 
Overall, \ours achieves a better balance for refusal behavior on adversarial questions. 

\begin{table}[tb]
\centering
\small
\caption{Performance of refusal on LoCoMo.
Among $1,986$ queries, $446$ are unanswerable. 
Precision is the proportion of predicted unanswerable cases that are indeed unanswerable (correct refusals as ``no information available''). 
Recall is the proportion of truly unanswerable cases correctly predicted as such. 
The F1 score is computed from these precision and recall values.}
\vspace{0.2em}
\label{tab:unanswerable-detection}
\begin{tabular}{l|rrrr}
\toprule
Methods & \# of Refusals & Precision (\%) & Recall (\%) & F1 (\%) \\
\midrule
Graphiti & $954$  & $38.9$ & $\mathbf{83.6}$ & $53.1$ \\
Mem0     &  $90$  & $40.0$ &  $8.1$ & $13.5$ \\
\ours                  & $344$  & $\mathbf{73.3}$ & $56.8$ & $\mathbf{64.0}$ \\
\bottomrule
\end{tabular}
\end{table}


\subsection{Human Evaluation}

To demonstrate the validity of using LLM as a judge in episodic memory, especially for reasoning tasks, we randomly selected $100$ samples from LoCoMo and conducted a human evaluation (Table \ref{tab:metrics_human_correlation}).
Both the LLM judge and the human evaluation use binary scores.
The LLM judge matched human scores in $93\%$ of cases, with only $7$ discrepancies.
Five LLM-accepted answers were manually judged as incorrect due to incomplete lists or temporal reasoning errors. Conversely, two LLM-rejected answers were manually judged as correct, because valid paraphrases were not recognized.
Though some limitations remain, these findings suggest that using LLM as a judge produces evaluations most closely aligned with human judgment compared to traditional metrics. 

\begin{table}[tb]
\vspace{-10pt}
\centering
\small
\caption{Comparison between automatic metrics and human evaluation. Mean denotes the mean value over $100$ selected LoCoMo samples. The correlations are compared with human evaluation.}
\vspace{0.2em}
\begin{tabular}{l|ccc}
\toprule
Metric & \textbf{Mean} & \textbf{Pearson $r$ w/ Human} & \textbf{Spearman $\rho$ w/ Human} \\ \toprule
Human Evaluation & $0.710$ & -- & -- \\
F1               & $0.410$ & $0.551$ & $0.603$ \\
BLEU-1           & $0.284$ & $0.417$ & $0.531$ \\
LLM Judge        & $0.740$ & $\mathbf{0.827}$ & $\mathbf{0.827}$ \\ 
\bottomrule
\end{tabular}
\label{tab:metrics_human_correlation}
\end{table}

\subsection{Error Analysis}

We conduct an error analysis for \ours on LoCoMo and Complex-TR.
In $100$ sampled LoCoMo errors, \textbf{the most common are selection or grounding errors ($46\%$)}, where \ours locates the correct or similar slot but assigns the wrong value or misinterprets the referent in details. For example, when asked ``\textit{What is Nate’s favorite video game?}'' \ours responded ``\textit{Catan}'', but that was only one of Nate's interests rather than his favorite ``\textit{Xenoblade Chronicles}''. 
\textbf{Temporal or numerical reasoning mistakes ($19\%$)} include errors in relative dates or durations. For instance, to the question ``\textit{When is Nate hosting a gaming party?}'' the model answered ``\textit{18–19 June 2022}'' while the gold reference was ``\textit{the weekend after 3 June 2022}''. 
\textbf{Another 18\% are abstentions despite the evidence being retrieved}, where the model incorrectly claims that no information is available, even though the gold answer is present.
In $100$ sampled Complex-TR errors, the most frequent failure mode ($42\%$) was \textbf{temporal window mismatch}, where correct entities are retrieved but misaligned with the specified time span. 
For instance, when asked ``\textit{Which employers did Ott-Heinrich Keller work for from Aug 1941 to Mar 1945?}'' the gold answer included both ``\textit{the Naval Academy at Mürwik}'' and ``\textit{the University of Münster}'', but our prediction listed only the first. 
Roughly $21\%$ of cases were due to \textbf{incomplete or inconsistent multi-entity lists}, where some retrieved items are missing or extraneous.
About $18\%$ are \textbf{offset direction mistakes}, such as confusing \textit{before} and \textit{after} or skipping to a later hop. For the question ``\textit{Which employer did Karyn A. Temple work for after RIAA?}'' where the gold answer was ``\textit{the U.S. Department of Justice}'' but the model instead gave ``\textit{the Copyright Office}''. 
A smaller fraction ($\approx5\%$) of ours incorrectly returns ``\textit{no information available}'' despite gold facts being present.

\subsection{Comparative Analysis: \ours vs. RAG}

We provide a few qualitative examples to compare \ours with a RAG system using NV-Embed-v2.
\ours outperforms NV-Embed-v2 on questions that require disambiguation across brand categories and reconciliation of time-stamped events, while the embedding baseline shines on straightforward interval calculations. 
For Q1 in Table \ref{tab:nvembed_comparison}, \ours extracts and reasons over a normalized gist, surfacing an explicit dated summary (``\textit{[19 Dec 2023] John got an amazing deal with a renowned outdoor gear company}''), and therefore matches the gold label exactly. NV-Embed-v2 selects a semantically similar but incorrect brand vertical (beverage). 
For Q2, \ours correctly chains events and compares timestamps (Susie was adopted around Aug 2021, while Seraphim was adopted last year), whereas the embedding method fixates on an earlier mention and ignores the newer adoption. 
Conversely, for the comparative Q3, NV-Embed-v2 retrieves both dates cleanly and computes the three-month interval, while \ours misaligns events in a longer context, yielding a one-month error. 
Overall, these examples indicate that time-aware gist extraction and agentic inference provide generally better robustness to distractors and temporal ambiguity, whereas a common RAG pipeline can be reliable when the answer reduces to simple computation over accurately retrieved facts. 
An additional comparative analysis between \ours and TISER on Complex-TR can be found in Appendix \ref{subsec:case studies}.

\begin{table}[tb]
\caption{Comparison between NV-Embed-v2 and \ours on LoCoMo. T or F indicates the LLM’s judgment (True or False).}
\vspace{0.2em}
\label{tab:nvembed_comparison}
\centering
\setlength{\tabcolsep}{3pt}
\footnotesize
\begin{adjustbox}{width=\linewidth}
\small
\begin{tabular}{p{0.32\linewidth} p{0.18\linewidth} p{0.22\linewidth} p{0.24\linewidth}}
\toprule
\textbf{Question} & \textbf{Gold Answer} & \textbf{NV-Embed-v2} & \textbf{\ours} \\
\midrule
Q1: What kind of deal did John get in December? & Deal with an outdoor gear company & Endorsement deal with a beverage company (F) &
Endorsement deal with an outdoor gear company (T) \\\midrule
Q2: Which pet did Jolene adopt more recently, Susie or Seraphim? &
Seraphim & Susie (F) & Seraphim (T) \\\midrule
Q3: How many months passed between Andrew adopting Toby and Buddy? &
Three months & 3 months (T) & 1 month (F) \\
\bottomrule
\end{tabular}
\end{adjustbox}
\end{table}

\section{Conclusion}
Challenges like episodic recollection and reasoning are far from solved for language agents.
We presented \ours, a time-aware episodic memory framework. 
The proposed hybrid memory graph unifies concept-level and context-level information with flexible temporal awareness, while the agentic retriever enables integration of retrieval and reasoning.
\ours consistently demonstrates better performance on episodic recollection and reasoning tasks across four benchmarks, with better refusal behavior for unanswerable queries and higher token efficiency.
\ours indicates a promising step toward more reliable long-horizon language agents.
Future work should consider long-term memory for language agents operating in more complex environments.
Building memories in a streaming format also poses an engineering challenge compared to offline batch indexing.


\subsubsection*{Acknowledgments}

The authors would like to thank colleagues from the OSU NLP group and Intuit AI Research for constructive discussions.
This material is based upon work supported by the National Science Foundation under Grant No. 2443149.
Any opinions, findings, and conclusions or recommendations expressed in this material are those of the author(s) and do not necessarily reflect the views of the National Science Foundation.
This work is also supported by an Alfred P. Sloan Foundation Fellowship.

\subsubsection*{Reproducibility statement}

We have provided the implementation details for reproducing the method and experiments.
All datasets used in this paper are publicly available, and we present dataset sampling in \S \ref{subsec:datasets}.
The metrics calculation is shown in \S \ref{subsec:metrics}.
The implementation details of \ours and the methods for comparison are presented in \S \ref{subsec:implementation-details}. 
The details on Mem0 and Graphiti are further provided in Appendix \ref{sec:appendix-implementation-details}.

\bibliography{iclr2026_conference}
\bibliographystyle{iclr2026_conference}

\clearpage
\appendix
\AppendixTOC

\section{The Use of Large Language Models}

Large language models did not play a significant role in the ideation or writing of this paper. They were only used for grammar checking and minor polishing of individual sentences.

\section{Formal Definition}

Let $\mathcal{M}=(V,E)$ be a typed multigraph with
$V = V_{\text{gist}} \cup V_{\text{phrase}}$ and
$E = E_{\text{rel}} \cup E_{\text{ctx}} \cup E_{\text{syn}}$.
The graph thus contains two node types and three edge types.

\paragraph{Node}
Each \textit{gist node} $g \in V_{\text{gist}}$ stores a natural-language episode summary $\text{text}(g)$ and an optional time scope $\tau(g)$ (point or interval), and represents a context-level, human-readable description of an event that jointly encodes its situational dimensions, such as participants, actions, objects, locations, intentions, and quantities, anchored to that time scope.
Each \textit{phrase node} $p \in V_{\text{phrase}}$ stores a short string $\text{name}(p)$ and represents a concept-level element extracted from a factual triple $(\text{subject}, \text{predicate}, \text{object})$, typically denoting a participant, action, or object in the event.
Phrase nodes can inherit temporal qualifiers such as \texttt{point\_in\_time}, \texttt{start\_time}, or \texttt{end\_time} from the underlying fact.

\paragraph{Edge}
Each \textit{relation edge} $e=(p_s, r, p_o, \tau(e)) \in E_{\text{rel}}$ links a subject phrase node $p_s$ and an object phrase node $p_o$ with a predicate $r$ and a validity interval $\tau(e)$, encoding fact-level relations such as ``Alice \text{---bought$\rightarrow$} cows'' together with their temporal scope derived from these qualifiers.
Each \textit{context edge} $e=(g,p) \in E_{\text{ctx}}$ connects a gist node $g$ to a phrase node $p$ extracted from the same source chunk that gave rise to $g$, thereby linking an abstract episode summary (e.g., ``Alice bought cows on Jan.\ 17th.'') to its underlying factual triple $(\text{Alice}, \text{bought}, \text{cows})$ and associated temporal qualifier (e.g., \texttt{point\_in\_time} = ``Jan.\ 17th'').
Each \textit{synonymy edge} $e=(g_i,g_j) \in E_{\text{syn}}$ links two gist nodes whose text embeddings exceed a similarity threshold, clustering semantically equivalent or highly related episodes and enriching the graph with associative links.

\paragraph{Index}
We maintain retrieval indices over this graph: an embedding index over $\text{text}(g)$ and $\text{name}(p)$, and a lexical (BM25) index over the surface forms of both gists and facts.

\paragraph{Algorithm} The algorithmic procedures for indexing and agentic inference are shown in Algorithm \ref{alg:index} and \ref{alg:react_infer}, respectively.

\begin{algorithm}[tb]
\caption{Indexing$(\mathcal{D}; P_{\text{gist}}, P_{\text{fact}}, \theta_{\text{syn}})$}
\label{alg:index}
\begin{algorithmic}[1]
\Require Event statements $\mathcal{D}=\{d_i\}$ with timestamps when available; prompts for gist and fact extraction $P_{\text{gist}}, P_{\text{fact}}$; synonymy threshold $\theta_{\text{syn}}$.
\Ensure Typed memory graph $\mathcal{M}$ and retrieval indices.
\State Initialize $V_{\text{gist}}, V_{\text{phrase}}, E_{\text{rel}}, E_{\text{ctx}}, E_{\text{syn}} \gets \emptyset$.
\For{$d \in \mathcal{D}$}
  \State $\mathcal{G} \gets \textsc{ExtractGists}(d; P_{\text{gist}})$ \Comment{LLM returns a set of gists with $\text{text}(g)$ and $\tau(g)$}
  \For{each $g \in \mathcal{G}$}
    \State add $g$ to $V_{\text{gist}}$
  \EndFor
  \State $\mathcal{F} \gets \textsc{ExtractFacts}(d; P_{\text{fact}})$ \Comment{LLM returns set of $(p_s,r,p_o,\tau)$}
  \State $P_d \gets \emptyset$ \Comment{phrase nodes appearing in this $d$}
  \For{each $(p_s,r,p_o,\tau) \in \mathcal{F}$}
    \State add $p_s,p_o$ to $V_{\text{phrase}}$ if new
    \State add $e=(p_s,r,p_o,\tau)$ to $E_{\text{rel}}$
    \State add $p_s,p_o$ to $P_d$
  \EndFor
  \For{each $g \in \mathcal{G}$}
    \For{each $p \in P_d$}
      \State add $(g,p)$ to $E_{\text{ctx}}$ \Comment{bind all gists to all phrases from the same $d$}
    \EndFor
  \EndFor
\EndFor
\ForAll{pairs $(u,v)$ in $V_{\text{gist}}$}
  \If{$\textsc{Sim}(u,v) \ge \theta_{\text{syn}}$}
    \State add $(u,v)$ to $E_{\text{syn}}$ 
  \EndIf
\EndFor
\State Build embedding/BM25 indices over $V$ and $\tau(\cdot)$
\State \Return $\mathcal{M}=(V,E)$ with indices
\end{algorithmic}
\end{algorithm}

\begin{algorithm}[tb]
\caption{AgenticInference$(q,\mathcal{M}, T_{\max})$}
\label{alg:react_infer}
\begin{algorithmic}[1]
\Require Query $q$; hybrid memory graph $\mathcal{M}$ with indices; iteration limit $T_{\max}$
\State $\mathcal{E} \gets \emptyset$ \Comment{evidence: accumulated gists and facts}
\State $\mathcal{H} \gets []$ \Comment{interaction history (thoughts, actions, observations)}
\State $k \gets 0$ \Comment{iteration counter}
\While{$k < T_{\max}$} 
  \State $t \gets \textsc{LLM\_Plan}(q, \mathcal{E}, \mathcal{H})$ \Comment{reasoning step}
  \State $a \gets \textsc{LLM\_SelectAction}(t)$
  \If{$a.\text{name}=\texttt{output\_answer}$}
      \State \Return $\textsc{LLM\_Synthesize}(q,\mathcal{E},\mathcal{H})$ \Comment{final answer based on gathered evidence}
  \ElsIf{$a.\text{name}=\texttt{semantic\_retrieve}$}
      \State $(S, O) \gets \textsc{SemanticRetrieve}(\mathcal{M}, a.\text{subquery})$ \Comment{embedding search; returns seed nodes $S$ and observations $O$}
      \State $\mathcal{E} \gets \mathcal{E} \cup O$; $\mathcal{H} \gets \mathcal{H} \,\|\, [t, a, O]$
  \ElsIf{$a.\text{name}=\texttt{lexical\_retrieve}$}
      \State $(S, O) \gets \textsc{LexicalRetrieve}(\mathcal{M}, a.\text{subquery})$ \Comment{BM25 search}
      \State $\mathcal{E} \gets \mathcal{E} \cup O$; $\mathcal{H} \gets \mathcal{H} \,\|\, [t, a, O]$
  \ElsIf{$a.\text{name}=\texttt{find\_gist\_contexts}$}
      \State $O \gets \textsc{FindGistContexts}(\mathcal{M}, a.\text{seeds}, a.\text{temporal\_constraints})$
      \State $\mathcal{E} \gets \mathcal{E} \cup O$; $\mathcal{H} \gets \mathcal{H} \,\|\, [t, a, O]$
  \ElsIf{$a.\text{name}=\texttt{find\_entity\_contexts}$}
      \State $O \gets \textsc{FindEntityContexts}(\mathcal{M}, a.\text{subject}, a.\text{predicate}, a.\text{object}, a.\text{temporal\_constraints})$ \Comment{filter by phrase/predicate/temporal constraints}
      \State $\mathcal{E} \gets \mathcal{E} \cup O$; $\mathcal{H} \gets \mathcal{H} \,\|\, [t, a, O]$
  \EndIf
  \State $k \gets k + 1$
\EndWhile
\State \Return $\textsc{LLM\_Synthesize}(q,\mathcal{E},\mathcal{H})$ 
\end{algorithmic}
\end{algorithm}

\section{Detailed Results}

\subsection{LoCoMo}
\label{subsec:results-on-locomo}

The LoCoMo performance is presented in Table \ref{tab:locomo_part1} and Table \ref{tab:locomo_part2}. 
For Oracle Message, while using annotations avoids retrieval challenges, QA remains a non-trivial task.
Overall, large embedding models serve as strong baselines, and structure-augmented memory methods show inferior performance. 
Compared with NV-Embed-v2, HippoRAG 2 achieves only comparable LLM-J scores ($+1.0$). 
In contrast, \ours-I outperforms NV-Embed-v2 on both F1 and J scores (F1 $+2.8$, J $+3.2$), achieving performance most closely aligned with the oracle messages among all evaluated methods.
\ours-S further improves the J score by $1.3$ and attains the highest J scores on LoCoMo. 
We find that \ours-S outperforms \ours-I in many single-session settings, suggesting that most queries do not require complex multi-step retrieval, which can instead introduce contextual noise. 
A notable bias of this benchmark is that multi-session queries account for only $14.2\%$ of all queries.
Importantly, unlike previous studies, we also evaluate performance under adversarial settings, which we regard as essential for evaluating a system’s refusal capability and mitigating hallucinations.
A more detailed analysis of model refusal behavior is provided in \S \ref{subsec:refusal-performance}.
Full-Context achieves the best results on multi-session questions, which may suggest that this benchmark imposes only limited demands on long-context reasoning ability.
Additional results on LoCoMo are also reported in Appendix \ref{subsec:results-on-locomo}.

By default, we use the top-10 chunks on LoCoMo for embedding models. Notably, we also evaluate NV-Embed-v2 with only the top-3 chunks (messages) and find that it constitutes a strong baseline. Even under this constrained setting, it surpasses Qwen3-Embedding-8B and some structure-augmented memory systems, except for HippoRAG 2, which is configured to use the top-3 sessions.

\begin{table}[tbh]
\centering
\small
\caption{Performance (\%) on LoCoMo (Part 1/2).}
\vspace{0.2em}
\label{tab:locomo_part1}
\resizebox{\textwidth}{!}{
\addtolength{\tabcolsep}{-3pt}
\begin{tabular}{l|ccc|ccc|ccc}
\toprule
 & \multicolumn{3}{c|}{Avg ($1{,}986$)} & \multicolumn{3}{c|}{Single-Hop ($321$)} & \multicolumn{3}{c}{Multi-Hop ($282$)} \\
 \cmidrule(lr){2-4} \cmidrule(lr){5-7} \cmidrule(lr){8-10}
Methods & F1 & BLEU-1 & LLM-J & F1 & BLEU-1 & LLM-J & F1 & BLEU-1 & LLM-J \\ \midrule
\rowcolor{gray!15}Oracle Message   & $48.0$ & $38.3$ & $81.0$ & $36.5$ & $28.2$ & $85.7$ & $41.6$ & $26.9$ & $84.8$ \\
\rowcolor{gray!15}Full-Context     & $37.8$ & $28.6$ & $76.7$ & $29.1$ & $20.6$ & $84.4$ & $34.2$ & $18.8$ & $75.2$ \\\midrule
\rowcolor{gray!8}
\multicolumn{10}{c}{\textit{\textbf{Large Embedding Models}}} \\
Qwen3-Embed-8B & $35.3$ & $28.9$ & $64.2$ & $24.8$ & $18.1$ & $64.2$ & $24.6$ & $15.1$ & $57.1$ \\
NV-Embed-v2 (7B)  & $39.6$ & $31.0$ & $73.0$ & $30.9$ & $\underline{22.6}$ & $77.9$ & $29.4$ & $17.8$ & $64.9$ \\
\ \ \ \ w/ Top-3 Messages & $39.0$ & $\underline{32.1}$ & $67.2$ & $26.5$ & $20.2$ & $68.2$ & $21.1$ & $13.6$ & $51.4$ \\ 
\rowcolor{gray!8}
\multicolumn{10}{c}{\textit{\textbf{Structure-Augmented Memory}}} \\
Mem0       & $25.1$ & $18.0$ & $49.7$ & $10.2$ & \ \ $6.3$ & $32.7$ & $29.9$ & $15.9$ & $52.1$ \\
Graphiti   & $33.7$ & $28.9$ & $52.5$ & \ \ $6.4$  & \ \ $3.4$ & $30.8$ & $25.1$ & $16.1$ & $53.6$ \\
HippoRAG 2 & $39.0$ & $30.8$ & $74.0$ & $25.8$ & $18.0$ & $72.0$ & $30.7$ & $\mathbf{19.5}$ & $\mathbf{72.3}$ \\
\rowcolor{gray!8}
\multicolumn{10}{c}{\textit{\textbf{Ours}}} \\
\ours-I        & $\mathbf{42.4}$ & $\mathbf{32.7}$ & $\underline{76.2}$ & $\mathbf{37.9}$ & $\mathbf{25.2}$ & $\underline{81.3}$ & $\mathbf{32.6}$ & $18.7$ & $\underline{70.2}$ \\
\ours-S & $\underline{41.3}$ & $31.5$ & $\mathbf{77.5}$ & $\underline{35.4}$ & $\underline{22.6}$ & $\mathbf{86.0}$ & $\underline{31.6}$ & $\underline{18.9}$ & $69.5$ \\
\bottomrule
\end{tabular}
}
\end{table}

\begin{table}[tbh]
\centering
\small
\caption{Performance (\%) on LoCoMo (Part 2/2). The adversarial number only represents refusal performance partially (see \S \ref{subsec:refusal-performance} for details).}
\vspace{0.2em}
\label{tab:locomo_part2}
\resizebox{\textwidth}{!}{
\addtolength{\tabcolsep}{-3pt}
\begin{tabular}{l|ccc|ccc|ccc}
\toprule
 & \multicolumn{3}{c|}{Open-Domain ($841$)} & \multicolumn{3}{c|}{Temporal ($96$)} & \multicolumn{3}{c}{Adversarial ($446$)} \\
 \cmidrule(lr){2-4} \cmidrule(lr){5-7} \cmidrule(lr){8-10}
Methods & F1 & BLEU-1 & LLM-J & F1 & BLEU-1 & LLM-J & F1 & BLEU-1 & LLM-J \\ \midrule
\rowcolor{gray!15}Oracle Message   & $43.8$ & $30.5$ & $86.0$ & $36.7$ & $23.7$ & $59.4$ & $70.6$ & $70.7$ & $70.6$ \\
\rowcolor{gray!15}Full-Context     & $36.0$ & $23.9$ & $88.3$ & $25.6$ & $15.9$ & $54.2$ & $52.2$ & $52.1$ & $55.0$ \\ \midrule
\rowcolor{gray!8}
\multicolumn{10}{c}{\textit{\textbf{Large Embedding Models}}} \\
Qwen3-Embed-8B & $27.9$ & $19.6$ & $67.3$ & $25.1$ & $15.3$ & $45.8$ & $65.9$ & $66.0$ & $66.8$ \\
NV-Embed-v2 (7B)  & $36.8$ & $24.6$ & $82.6$ & $26.2$ & $16.2$ & $51.0$ & $60.8$ & $60.8$ & $61.2$ \\
\ \ \ \ w/ Top-3 Messages & $33.7$ & $23.3$ & $71.7$ & $23.1$ & $13.1$ & $43.8$ & $\underline{72.9}$ & $\underline{72.9}$ & $\underline{73.1}$ \\
\rowcolor{gray!8}
\multicolumn{10}{c}{\textit{\textbf{Structure-Augmented Memory}}} \\
Mem0       & $\underline{38.2}$ & $\mathbf{27.5}$ & $57.2$ & $24.1$ & $15.7$ & $46.9$ & \ \ $8.3$  & $10.4$ & $46.6$ \\
Graphiti   & $21.9$ & $15.6$ & $45.0$ & $23.0$ & $14.6$ & $45.8$ & $\mathbf{83.2}$ & $\mathbf{83.4}$ & $\mathbf{83.0}$ \\
HippoRAG 2 & $35.5$ & $23.9$ & $81.7$ & $\mathbf{28.8}$ & $\mathbf{18.5}$ & $\underline{56.3}$ & $62.6$ & $62.6$ & $65.7$ \\
\rowcolor{gray!8}
\multicolumn{10}{c}{\textit{\textbf{Ours}}} \\
\ours-I & $\mathbf{39.1}$ & $\underline{26.4}$ & $\underline{83.5}$ & $25.5$ & $\underline{17.5}$ & $\underline{56.3}$ & $61.7$ & $62.0$ & $66.8$ \\
\ours-S & $37.3$ & $24.6$ & $\mathbf{85.1}$ & $\underline{28.2}$ & $\mathbf{18.5}$ & $\mathbf{63.5}$ & $62.1$ & $61.8$ & $65.3$ \\
\bottomrule
\end{tabular}
}
\end{table}

\subsection{REALTALK}
\label{subsec:results-on-realtalk}

Overall, REALTALK proves to be more challenging than LoCoMo, as evidenced by the consistently lower performance of all methods. Constructed from real human interactions, it naturally contains more noise and casual expressions.
The overall performance distribution is similar to that observed on LoCoMo. NV-Embed-v2 remains a strong baseline as an embedding model, while structure-enhanced memory methods fall short in comparison.
\ours is the only one that surpasses the Full-Context method, particularly on tasks involving temporal reasoning.

\begin{table}[tbh]
\centering
\small
\caption{Performance (\%) on REALTALK.}
\vspace{0.2em}
\label{tab:perf_scaled_no_em}
\resizebox{\textwidth}{!}{
\addtolength{\tabcolsep}{-4pt}
\begin{tabular}{l|ccc|ccc|ccc|ccc}
\toprule
& \multicolumn{3}{c|}{Overall ($728$)} & \multicolumn{3}{c|}{Multi-hop ($301$)} & \multicolumn{3}{c|}{Commonsense ($108$)} & \multicolumn{3}{c}{Temporal ($319$)} \\
 \cmidrule(lr){2-4} \cmidrule(lr){5-7} \cmidrule(lr){8-10} \cmidrule(lr){11-13}
Methods & F1 & BLEU-1 & LLM-J & F1 & BLEU-1 & LLM-J & F1 & BLEU-1 & LLM-J & F1 & BLEU-1 & LLM-J \\ \midrule
\rowcolor{gray!15}Full-Context & $25.3$ & $18.6$ & $65.1$ & $27.8$ & $21.7$ & $60.1$ & $19.5$ & $15.1$ & $55.6$ & $24.9$ & $16.8$ & $73.0$ \\\midrule
\rowcolor{gray!8}
\multicolumn{13}{c}{\textit{\textbf{Large Embedding Models}}} \\
Qwen3-Embed-8B & $20.2$ & $14.9$ & $52.5$ & $19.6$ & $16.2$ & $41.5$ & $14.5$ & $11.8$ & $39.8$ & $22.7$ & $14.8$ & $67.1$ \\
NV-Embed-v2 (7B) & $23.8$ & $17.7$ & $59.5$ & $24.5$ & $19.4$ & $51.2$ & $16.6$ & $13.0$ & $48.2$ & $25.6$ & $\mathbf{17.7}$ & $71.2$ \\
\rowcolor{gray!8}
\multicolumn{13}{c}{\textit{\textbf{Structure-Augmented Memory}}} \\
Mem0 & \ \ $9.8$ & \ \ $7.2$ & $14.3$ & $14.0$ & \ \ $9.6$ & $20.9$ & $12.0$ & \ \ $9.0$ & $22.2$ & \ \ $5.1$ & \ \ $4.3$ & \ \ $5.3$ \\
Graphiti & $15.1$ & $11.5$ & $35.3$ & $19.6$ & $15.5$ & $39.5$ & $\underline{16.7}$ & $13.2$ & $44.4$ & $10.5$ & \ \ $7.1$ & $28.2$ \\
HippoRAG 2 & $21.9$ & $16.2$ & $55.8$ & $26.1$ & $20.9$ & $51.8$ & $\mathbf{18.2}$ & $\underline{13.7}$ & $\mathbf{54.6}$ & $19.2$ & $12.6$ & $59.9$ \\
\rowcolor{gray!8}
\multicolumn{13}{c}{\textit{\textbf{Ours}}} \\
\ours-I & $\underline{25.6}$ & $\underline{18.1}$ & $\underline{63.7}$ & $\underline{26.6}$ & $\underline{21.6}$ & $\underline{55.8}$ & $16.1$ & $13.0$ & $45.4$ & $\mathbf{27.9}$ & $16.5$ & $\mathbf{77.4}$ \\
\ours-S & $\mathbf{26.2}$ & $\mathbf{19.2}$ & $\mathbf{65.3}$ & $\mathbf{28.6}$ & $\mathbf{23.1}$ & $\mathbf{58.8}$ & $\mathbf{18.2}$ & $\mathbf{14.7}$ & $\underline{53.7}$ & $\underline{26.7}$ & $\underline{17.1}$ & $\underline{75.2}$ \\
\bottomrule
\end{tabular}
}
\end{table}

\subsection{Complex-TR}
\label{subsec:results-on-complex-tr}

The performance on Complex-TR is shown in Table \ref{tab:complex-tr}.
On Complex-TR, embedding models remain strong baselines. 
Among structure-augmented memory methods, HippoRAG 2 slightly outperforms the embedding model (J $+1.1$), likely due to its alignment with the entity-centric nature of this dataset. 
Compared to \ours, the only competitive method is NV-Embed-v2 w/ TISER, which employs multi-step temporal reasoning prompts and even outperforms \ours-I on the time-to-event task (J $+1.4$), but falls short on average (J $-1.3$).
Compared with the event-to-event type, the time-to-event type requires more direct temporal resolution, which better suits TISER.

\ours primarily focuses on information extraction and agentic retrieval. 
For the reasoning component, we employ straightforward prompts (Appendix \ref{sec:prompts}) that leverage the context to answer questions. 
When we adopt TISER as the reasoning prompt for the final step, \ours w/ TISER achieves further improvements and delivers the highest performance (F1 $+7.3$, J $+2.4$, compared to \ours) overall.
Additionally, for such complex reasoning tasks, \ours-S lacks multi-hop reasoning capabilities (J $-6.0$), which demonstrates the necessity of employing agentic retrieval for autonomous reasoning.

\begin{table}[tbh]
\centering
\small
\caption{Performance (\%) on Complex-TR.}
\vspace{0.2em}
\label{tab:complex-tr}
\resizebox{\textwidth}{!}{
\addtolength{\tabcolsep}{-3pt}
\begin{tabular}{l|ccc|ccc|ccc}
\toprule
 & \multicolumn{3}{c|}{Avg} & \multicolumn{3}{c|}{Time to Event ($543$)} & \multicolumn{3}{c}{Event to Event ($457$)} \\
\cmidrule(lr){2-4} \cmidrule(lr){5-7} \cmidrule(lr){8-10}
Methods & F1 & BLEU-1 & LLM-J & F1 & BLEU-1 & LLM-J & F1 & BLEU-1 & LLM-J \\
\midrule
\rowcolor{gray!15}Full-context & $74.2$ & $68.0$ & $81.6$ & $73.0$ & $65.6$ & $80.3$ & $75.6$ & $70.8$ & $83.2$ \\  \midrule
\rowcolor{gray!8}
\multicolumn{10}{c}{\textit{\textbf{Large Embedding Models}}} \\
Qwen3-Embed-8B & $77.1$ & $71.4$ & $80.9$ & $74.5$ & $68.8$ & $79.0$ & $80.2$ & $74.5$ & $83.2$ \\
NV-Embed-v2 (7B) & $77.5$ & $71.9$ & $80.4$ & $77.0$ & $70.8$	& $79.2$ & $78.1$	& $73.2$	& $81.8$ \\
\ \ \ \ w/ TISER & $\underline{88.1}$ & $\underline{83.6}$ & $88.3$ & $\underline{88.9}$ & $\underline{83.5}$ & $\underline{90.4}$ & $\underline{87.1}$ & $\underline{83.7}$ & $85.8$ \\
\rowcolor{gray!8}
\multicolumn{10}{c}{\textit{\textbf{Structure-Augmented Memory}}} \\
Mem0 & $43.1$ & $35.1$ & $41.0$ & $43.5$ & $35.1$ & $40.9$ & $42.6$ & $35.1$ & $41.1$ \\
Graphiti & $76.6$ & $71.4$ & $78.8$ & $77.3$ & $71.1$ & $80.3$ & $75.6$ & $71.7$ & $77.0$ \\
HippoRAG 2 & $78.2$ & $72.7$ & $81.5$ & $78.0$ & $71.4$ & $81.8$ & $78.5$ & $74.4$ & $81.2$ \\
\rowcolor{gray!8}
\multicolumn{10}{c}{\textit{\textbf{Ours}}} \\
\ours-I & $83.3$ & $77.6$ & $\underline{89.6}$ & $80.3$ & $73.2$ & $89.0$ & $86.9$ & $82.8$ & $\mathbf{90.4}$ \\
\ \ \ \ w/ TISER & $\mathbf{90.6}$ & $\mathbf{86.0}$ & $\mathbf{92.0}$ & $\mathbf{92.5}$ & $\mathbf{87.1}$ & $\mathbf{94.3}$ & $\mathbf{88.1}$ & $\mathbf{83.8}$ & $\underline{89.3}$ \\
\ours-S & $78.5$ & $72.7$ & $82.6$ & $78.6$ & $71.6$ & $84.2$ & $78.5$ & $74.0$ & $80.7$ \\
\bottomrule
\end{tabular}
}
\end{table}

\subsection{Test of Time (Semantic)}
\label{subsec:results-on-tot}

The performance on the Test of Time is shown in Table \ref{tab:test of time}.
Here, Full-Context directly uses the original prompt from the dataset, which contains hundreds of facts per question on average.
Since Test of Time employs anonymous entity and relation labels, we also use BM25 as the retriever in addition to the embedding model. 
The results show that even the improved GPT-5-chat still falls short as a Full-Context method when compared with its predecessor, the economical GPT-4.1-mini (EM $+4.8$).
HippoRAG 2 heavily relies on graph traversal, and due to its lack of understanding of complex logic, its performance falls short of even embedding models.
TISER continues to serve as a strong baseline, while \ours is the only method that surpasses $90\%$, achieving an $+8.2$ EM improvement over Full-Context with TISER.
Beyond Table \ref{tab:test of time}, we further observe that \ours-I, when paired with stronger GPT-5-chat, can address nearly all eight types of challenges in this benchmark (EM $99.0$), underscoring the rationality of the agentic retrieval design.

\begin{table}[tbh]
\centering
\caption{Exact match (\%) on Test of Time (semantic). GPT-4.1-mini is the default LLM. While Full-Context adopts the full prompt from the original dataset, BM25 and NV-Embed-v2 provide shortened prompts by retrieving the top-$k$ facts.
BA: Before-after. ET: Event at time \textit{t}. EW: Event at what time. FL: First-last. EA: Event at the time of another event. NE: Number of events in time interval. RD: Relation duration. TL: Timeline. Each category contains $350$ samples. * denotes using a 160-sample subset as an approximation due to the high cost of full-dataset evaluation.}
\vspace{0.2em}
\label{tab:test of time}
\begin{adjustbox}{width=\textwidth,center}
\begin{tabular}{l|r|rrrrrrrr}
\toprule
Methods & Avg & BA & ET & EW & FL & EA & NE & RD & TL \\
\midrule
Full-Context (GPT-4.1-mini) & $79.7$ & $72.0$ & $86.9$ & $98.0$ & $81.1$ & $83.7$ & $76.3$ & $93.4$ & $46.3$ \\
\ \ \ \ w/ TISER & $84.9$ & $88.3$ & $91.4$ & $97.7$ & $89.1$ & $90.6$ & $64.0$ & $94.6$ & $63.4$ \\
Full-Context (GPT-5-chat) & $84.5$ & $80.0$ & $90.0$ & $99.4$ & $84.9$ & $91.1$ & $90.0$ & $82.3$ & $58.0$ \\\midrule
BM25 & $80.8$ & $71.1$ & $77.7$ & $\underline{98.6}$ & $\underline{92.3}$ & $64.3$ & $65.1$ & $98.3$ & $79.1$ \\
Qwen-Embed-8B & $70.3$ & $53.1$ & $88.3$ & $\mathbf{100.0}$ & $72.6$ & $42.9$ & $\underline{71.7}$ & $98.0$ & $35.7$ \\
NV-Embed-v2 (7B) & $68.9$ & $43.1$ & $86.6$ & $\mathbf{100.0}$ & $74.6$ & $48.0$ & $\mathbf{75.7}$ & $96.9$ & $26.6$ \\
\ \ \ \ w/ TISER & $68.9$ & $48.0$ & $86.0$ & $\mathbf{100.0}$ & $74.6$ & $51.7$ & $67.4$ & $96.9$ & $26.3$ \\
HippoRAG 2* & $66.9$ & $60.0$ & $\underline{90.0}$ & $\mathbf{100.0}$ & $75.0$ & $45.0$ & $50.0$ & $\mathbf{100.0}$ & $15.0$ \\
 \midrule
\ours-I & $\mathbf{93.1}$ & $\underline{97.4}$ & $\mathbf{99.4}$ & $\mathbf{100.0}$ & $92.0$ & $\mathbf{97.4}$ & $71.4$ & $94.0$ & $\mathbf{92.9}$ \\
\ \ \ \ w/ TISER & $\underline{90.6}$ & $\mathbf{98.6}$ & $\mathbf{99.4}$ & $\underline{98.6}$ & $\mathbf{95.4}$ & $\underline{93.7}$ & $68.3$ & $85.1$ & $\underline{85.4}$ \\
\ours-S & $72.5$ & $74.6$ & $78.0$ & $\mathbf{100.0}$ & $90.9$ & $64.3$ & $52.0$ & $\underline{99.7}$ & $20.9$ \\ 
\bottomrule
\end{tabular}
\end{adjustbox}
\end{table}

\subsection{Semantic Memory Capability}

Although episodic memory is a crucial aspect of long-term memory and the main focus of this paper, semantic memory should not be overlooked as another aspect. 
We select MuSiQue and 2Wiki from the benchmarks used in HippoRAG 2 to evaluate this capability.
From Table 7, we observe that Mem0 performs substantially worse than both NV-Embed-v2 and \ours across all metrics, with results nearly collapsing. 
This performance highlights limitations in Mem0’s memory construction process: a major part of the extracted facts is not added to its memory, and its use of a fixed dialog extraction pipeline is poorly aligned with the structure of passages that incorporate world knowledge. 
Mem0 frequently extracts incomplete or decontextualized sentences in which even the subject is ambiguous, e.g,. ``\textit{Managed La Liga clubs including Barcelona, Atlético Bilbao, Atlético Madrid, and Real Zaragoza}'' is extracted from a paragraph titled ``\textit{Ferdinand Daučík}'', and then this extracted sentence is not added or updated to the memory.
For comparison, though our method primarily focuses on episodic memory and reasoning, it still achieves performance comparable to a strong embedding model on this task, demonstrating its strong extensibility.

\begin{table}[tbh]
\centering
\small
\caption{Performance (\%) on QA tasks with semantic memory.}
\vspace{0.2em}
\begin{tabular}{l|rrr|rrr|rrr}
\toprule
 & \multicolumn{3}{c|}{Avg} & \multicolumn{3}{c|}{MuSiQue ($1,000$)} & \multicolumn{3}{c}{2Wiki ($1,000$)} \\ \midrule
Methods & F1 & BLEU-1 & LLM-J & F1 & BLEU-1 & LLM-J & F1 & BLEU-1 & LLM-J \\ \midrule
NV-Embed-v2 & $38.2$ & $30.5$ & $\mathbf{57.5}$ & $37.5$ & $31.9$ & $\mathbf{56.3}$ & $38.8$ & $29.1$ & $\mathbf{58.6}$ \\
Mem0 & $6.9$ & $5.1$ & $8.0$ & $7.4$ & $5.7$ & $9.2$ & $6.3$ & $4.5$ & $6.8$ \\
\ours & $\mathbf{38.2}$ & $\mathbf{32.1}$ & $55.3$ & $\mathbf{37.9}$ & $\mathbf{33.5}$ & $53.2$ & $\mathbf{38.6}$ & $\mathbf{30.8}$ & $57.4$ \\
\bottomrule
\end{tabular}
\label{tab:semantic memory}
\end{table}

\subsection{Ablation Study}
\label{subsec:appendix-ablation-study}

More detailed results of the ablation study are presented in Table \ref{tab:locomo-ablation} and Table \ref{tab:complex-tr-ablation}.
Overall, gists provide the primary context (J $-27.3\%$ on LoCoMo and $-8.7\%$ on Complex-TR), while facts offer indispensable supplementary support.
In particular, for LoCoMo, \ours-I w/o Facts achieved comparable performance to \ours across most tasks, but showed a clear drop in multi-hop questions (J $-7.1\%$). This highlights the crucial role of phrase nodes in bridging concepts across sessions and facilitating effective graph exploration.

\begin{table}[tbh]
\centering
\small
\caption{Ablation study on LoCoMo.}
\vspace{0.2em}
\addtolength{\tabcolsep}{-3pt}
\begin{tabular}{l|cc|cc|cc|cc|cc|cc}
\toprule
 & \multicolumn{2}{c|}{Avg} & \multicolumn{2}{c|}{Single-Hop} & \multicolumn{2}{c|}{Multi-Hop} & \multicolumn{2}{c|}{Open-Domain} & \multicolumn{2}{c|}{Temporal} & \multicolumn{2}{c}{Adversarial} \\
 \cmidrule(lr){2-3} \cmidrule(lr){4-5} \cmidrule(lr){6-7} \cmidrule(lr){8-9} \cmidrule(lr){10-11} \cmidrule(lr){12-13}
Methods & F1 & LLM-J & F1 & LLM-J & F1 & LLM-J & F1 & LLM-J & F1 & LLM-J & F1 & LLM-J \\ \midrule
\ours-I & $\mathbf{42.4}$  & $\underline{76.2}$ & $\underline{37.9}$ & $81.3$ & $\mathbf{32.6}$ & $\mathbf{70.2}$ & $\underline{39.1}$ & $83.5$ & $25.5$ & $\underline{56.3}$ & $61.7$ & $\underline{66.8}$ \\
\ \ \ \ w/o Gists & $31.7$ & $48.9$ & $19.7$ & $69.8$ & $20.5$ & $42.2$ & $12.0$ & $24.9$  & $14.5$ & $28.1$ & $\mathbf{88.3}$ & $\mathbf{88.1}$ \\
\ \ \ \ w/o Facts & $\underline{42.0}$ & $74.1$ & $\mathbf{45.7}$ & $\underline{81.9}$ & $28.2$ & $63.1$ & $\mathbf{41.7}$ & $\underline{84.0}$ & $\underline{25.6}$ & $54.2$ & $52.2$ & $61.0$ \\
\ours-S & $41.3$ & $\mathbf{77.5}$ & $35.4$ & $\mathbf{86.0}$ & $\underline{31.6}$ & $\underline{69.5}$ & $37.3$ & $\mathbf{85.1}$ & $\mathbf{28.2}$ & $\mathbf{63.5}$ & $\underline{62.1}$ & $65.3$ \\ 
\bottomrule
\end{tabular}
\label{tab:locomo-ablation}
\end{table}

\begin{table}[tbh]
\centering
\small
\caption{Ablation study on Complex-TR.}
\vspace{0.2em}
\label{tab:complex-tr-ablation}
\begin{tabular}{l|ccc|ccc|ccc}
\toprule
 & \multicolumn{3}{c|}{Avg} & \multicolumn{3}{c|}{Time to Event ($543$)} & \multicolumn{3}{c}{Event to Event ($457$)} \\
\cmidrule(lr){2-4} \cmidrule(lr){5-7} \cmidrule(lr){8-10}
Methods & F1 & BLEU-1 & LLM-J & F1 & BLEU-1 & LLM-J & F1 & BLEU-1 & LLM-J \\
\midrule
\ours-I & $\mathbf{83.3}$ & $\mathbf{77.6}$ & $\mathbf{89.6}$ & $\mathbf{80.3}$ & $\mathbf{73.2}$ & $\mathbf{89.0}$ & $\underline{86.9}$ & $\underline{82.8}$ & $\mathbf{90.4}$ \\
\ \ \ \ w/o Gists & $80.3$ & $\underline{75.9}$ & $80.9$ & $73.7$ & $69.1$ & $75.3$ & $\mathbf{88.1}$ & $\mathbf{84.0}$ &  $87.5$ \\
\ \ \ \ w/o Facts &  $\underline{80.5}$ & $74.5$ & $\underline{87.2}$ & $\underline{79.3}$ & $\underline{72.7}$ & $\underline{86.6}$ & $81.9$ & $76.6$ & $\underline{88.0}$  \\  
\ours-S & $78.5$ & $72.7$ & $82.6$ & $78.6$ & $71.6$ & $84.2$ & $78.5$ & $74.0$ & $80.7$ \\
\bottomrule
\end{tabular}
\end{table}

\subsection{Performance by Temporal Category}

\begin{table}[tbh]
\centering
\small
\caption{Our defined temporal categories on LoCoMo.}
\begin{tabular}{l r p{4.6cm} p{4.6cm}}
\toprule
\textbf{Category} & \textbf{Count} & \textbf{Description} & \textbf{Query Example} \\
\midrule
Existence Check &
$14$ &
Check for the presence of temporal facts, given a specific time frame &
Did Andrew have a pet dog during March 2023? \\
Event Timing &
$298$ &
Determine the specific time points or intervals when an event occurred &
When did Melanie paint a sunrise? \\
Event Attributes &
$302$ &
Identify attributes or characteristics of an event, e.g., location, participants, etc. &
What are some changes Caroline has faced during her transition journey? \\
Order &
$22$ &
Understand temporal relations between events, e.g., sequence, concurrency, overlap &
What did Melanie do after the road trip to relax? \\
Duration &
$44$ &
Determine the duration of an event or the time interval between two events &
How long has Caroline had her current group of friends for? \\
Aggregation &
$30$ &
Count occurrences of an event within a time frame, or count specific attributes of events &
How many times has Melanie gone to the beach in 2023? \\
Other &
$6$ &
Other temporal reasoning tasks not covered by the above categories &
Would Caroline want to move back to her home country soon? \\
Non-temporal &
$1270$ &
No temporal reasoning is required, but other situational elements&
Where did Oliver hide his bone once? \\
\bottomrule
\end{tabular}
\label{tab:temporal_categories}
\end{table}

We present performance metrics by temporal category to provide more analysis for temporal and non-temporal questions.
We set a few temporal categories, as shown in Table \ref{tab:temporal_categories}, and instruct GPT-4.1-mini to classify each query into one of the categories.

Then, we report REMem-I’s performance on LoCoMo in Table \ref{tab:locomo_temporal} according to the above categories, where `temporal` is the average value of all temporal categories, and `overall` is the average value of `temporal` and `none` categories. 
These results suggest that the performance of REMem-I on non-temporal questions is on par with the temporal ones.

\begin{table}[t]
\small
\caption{The performance on LoCoMo by temporal categories.}
\centering
\begin{tabular}{lrrrr}
\toprule
Temporal category & \# of samples & F1 & BLEU-1 & LLM-J \\
\midrule
Overall           & $1986$ & $42.4$ & $32.7$ & $76.2$ \\ \midrule
Non-temporal              & $1086$ & $41.8$ & $32.4$ & $75.0$ \\
Temporal          & $900$  & $43.2$ & $33.1$ & $77.7$ \\ \midrule
Existence Check  & $15$   & $59.5$ & $11.9$ & $86.7$ \\
Event Attributes & $450$  & $45.9$ & $37.9$ & $76.7$ \\
Event Timing     & $298$  & $39.1$ & $26.0$ & $82.6$ \\
Aggregation       & $35$   & $39.5$ & $33.9$ & $60.0$ \\
Duration          & $50$   & $35.2$ & $26.8$ & $72.0$ \\
Order             & $49$   & $48.4$ & $45.0$ & $71.4$ \\
Other             & $3$    & $42.4$ & $37.8$ & $100.0$ \\
\bottomrule
\end{tabular}
\label{tab:locomo_temporal}
\end{table}

\section{Prompts}
\label{sec:prompts}

\begin{figure}[htb]
    \centering
    \includegraphics[width=0.95\linewidth]{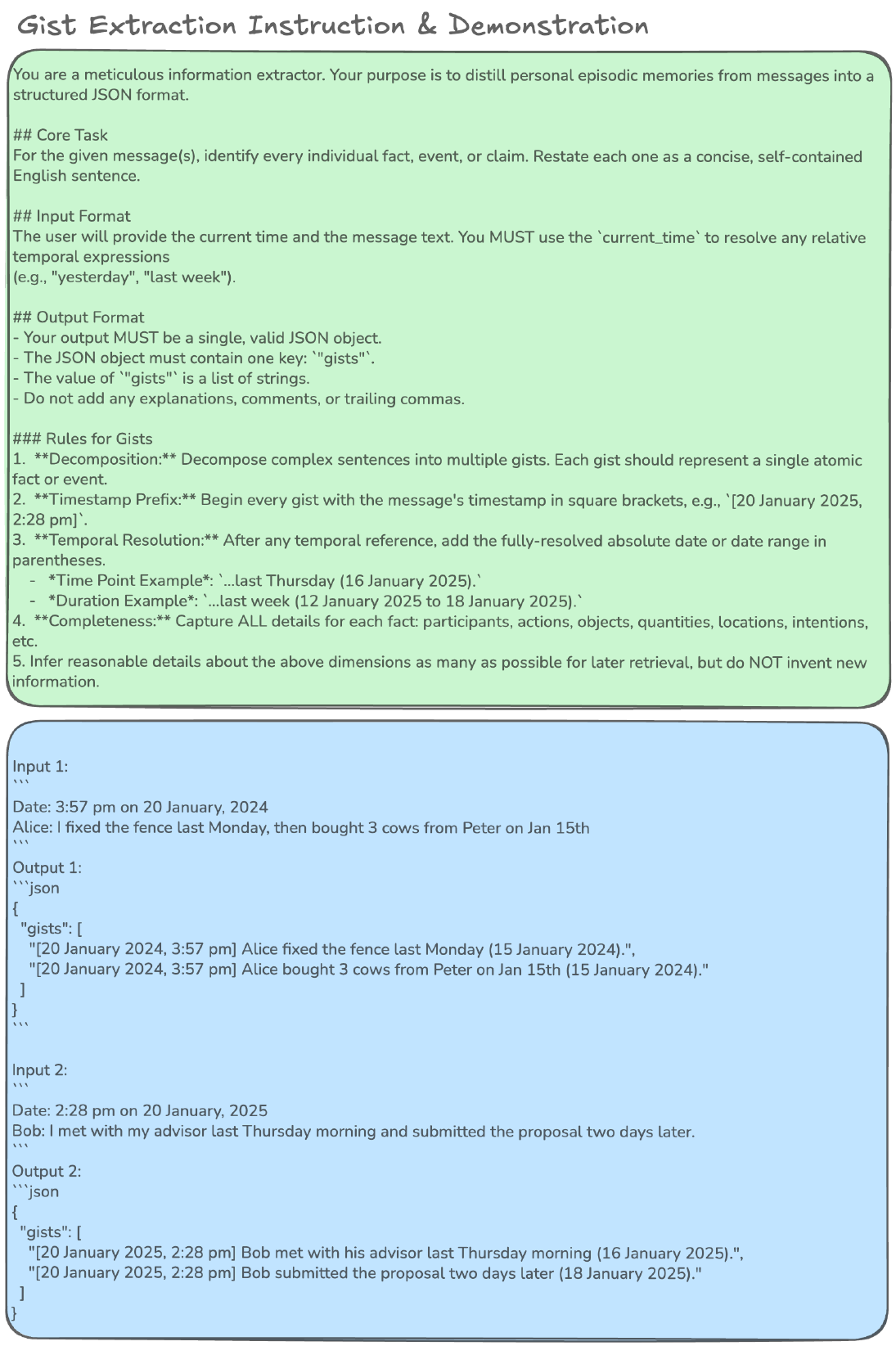}
    \caption{The prompts for gist extraction. The instructions and demonstrations are marked in different colors.}
    \label{fig:gist extraction}
\end{figure}

\begin{figure}[th]
    \centering
    \includegraphics[width=0.95\linewidth]{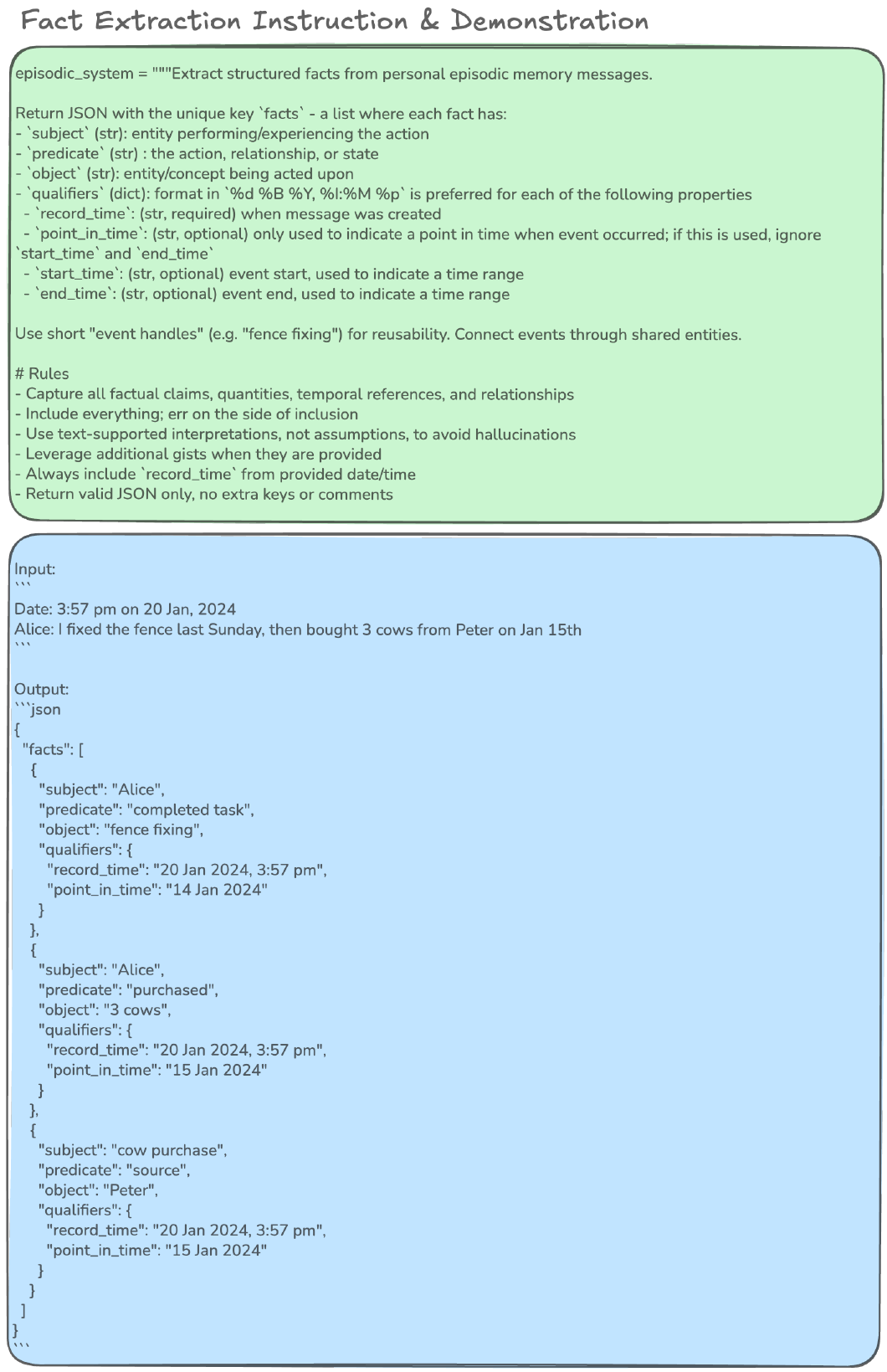}
    \caption{The prompts for fact extraction. The instructions and demonstrations are marked in different colors.}
    \label{fig:fact extraction}
\end{figure}

\begin{figure}[th]
    \centering
    \includegraphics[width=0.95\linewidth]{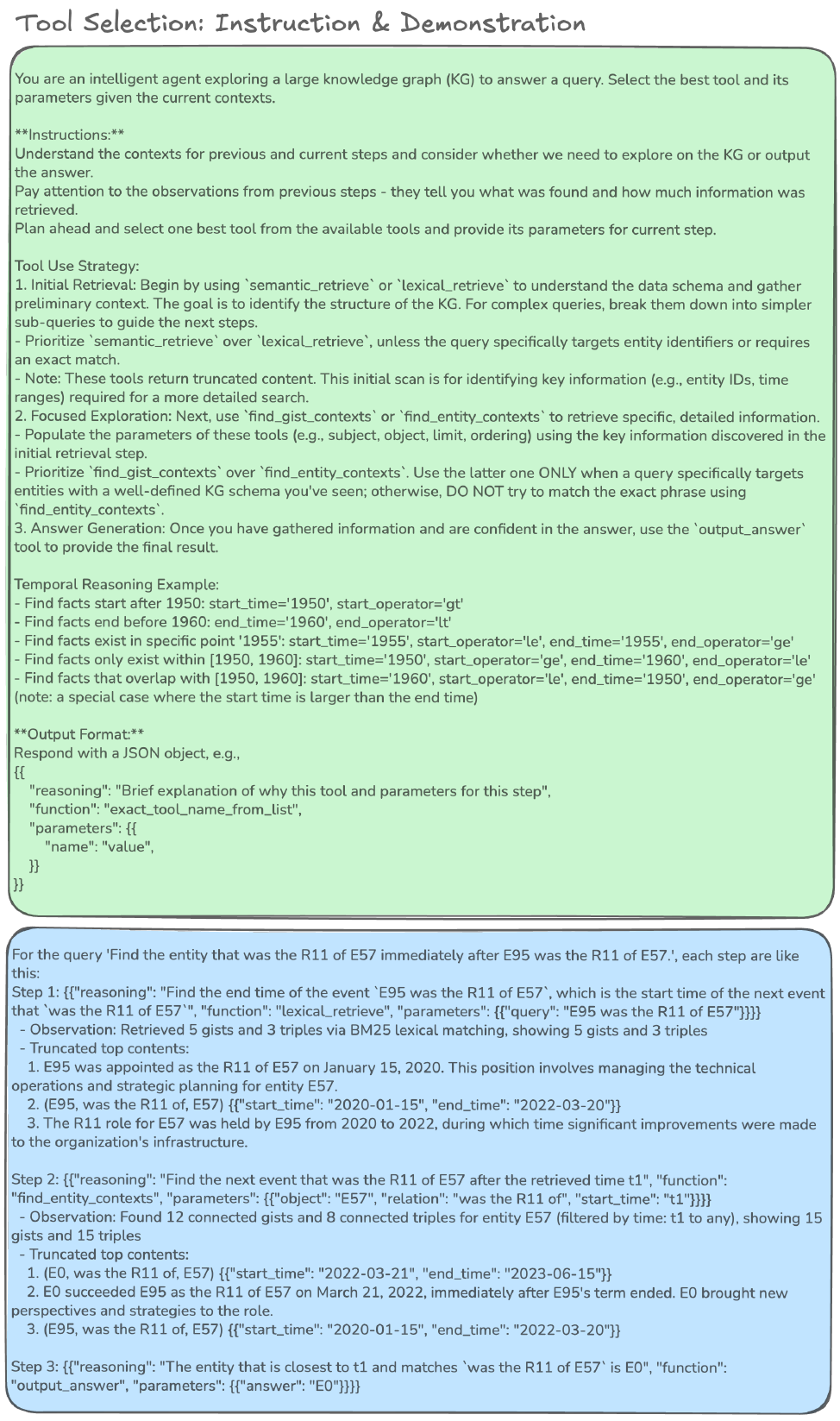}
    \caption{The prompts for tool selection. The instructions and demonstrations are marked in different colors.}
    \label{fig:tool selection}
\end{figure}

\begin{figure}[th]
    \centering
    \includegraphics[width=0.95\linewidth]{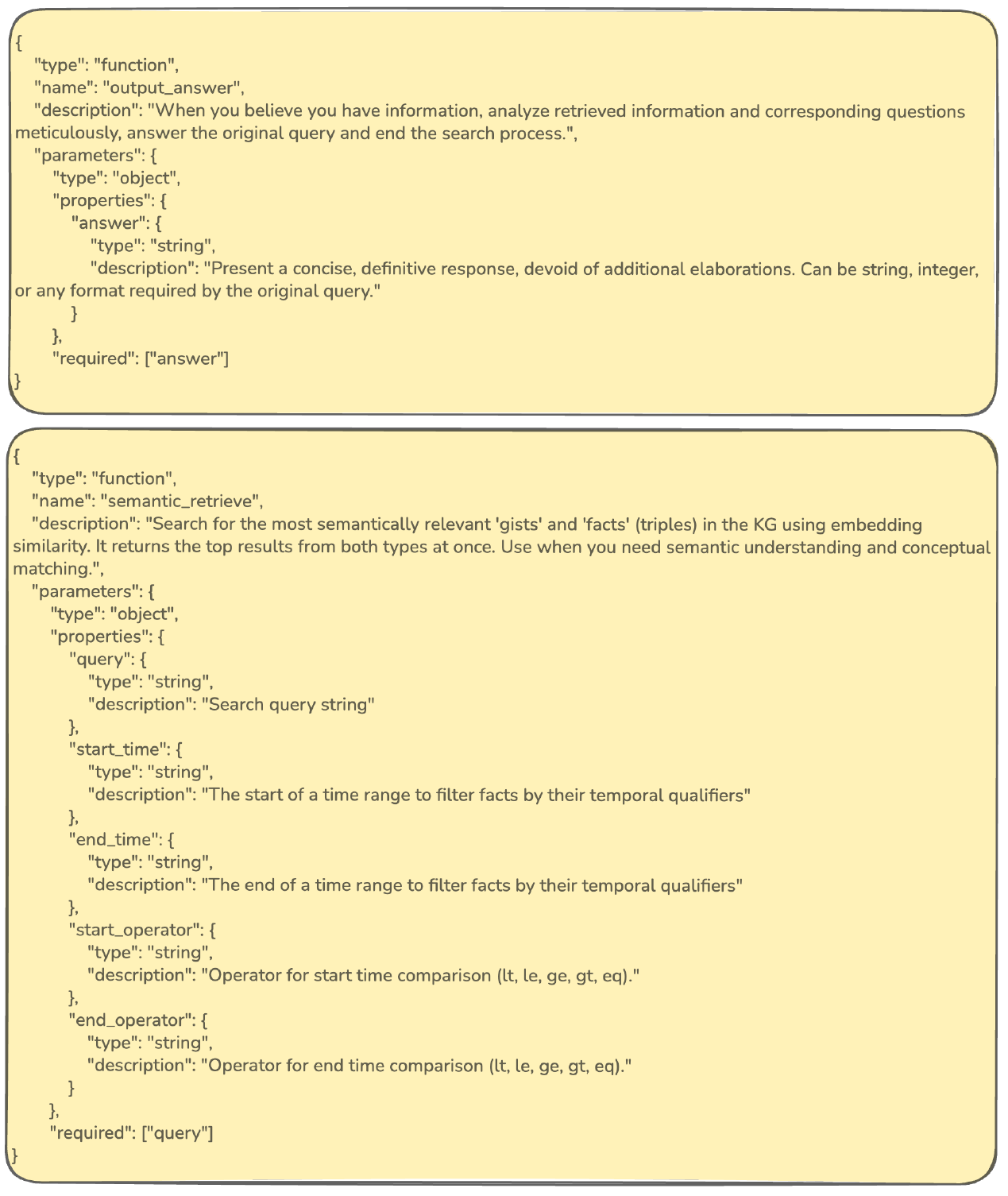}
    \caption{The tool descriptions for \textit{output\_answer} and \textit{semantic\_retrieve}.}
    \label{fig:tool description 1}
\end{figure}

\begin{figure}[th]
    \centering
    \includegraphics[width=0.95\linewidth]{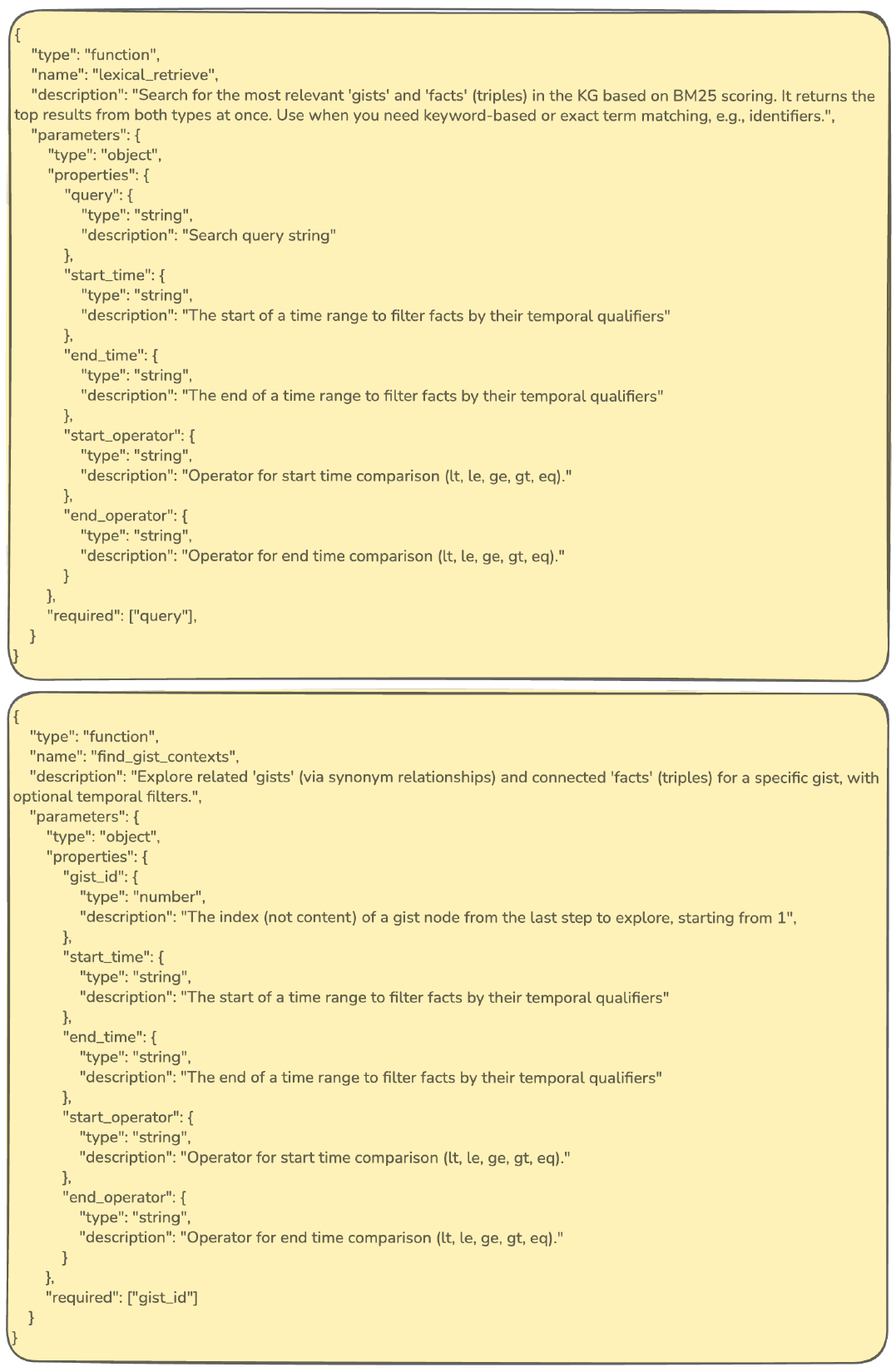}
    \caption{The tool descriptions for \textit{lexical\_retrieve} and \textit{find\_gist\_contexts}.}
    \label{fig:tool description 2}
\end{figure}

\begin{figure}[th]
    \centering
    \includegraphics[width=0.95\linewidth]{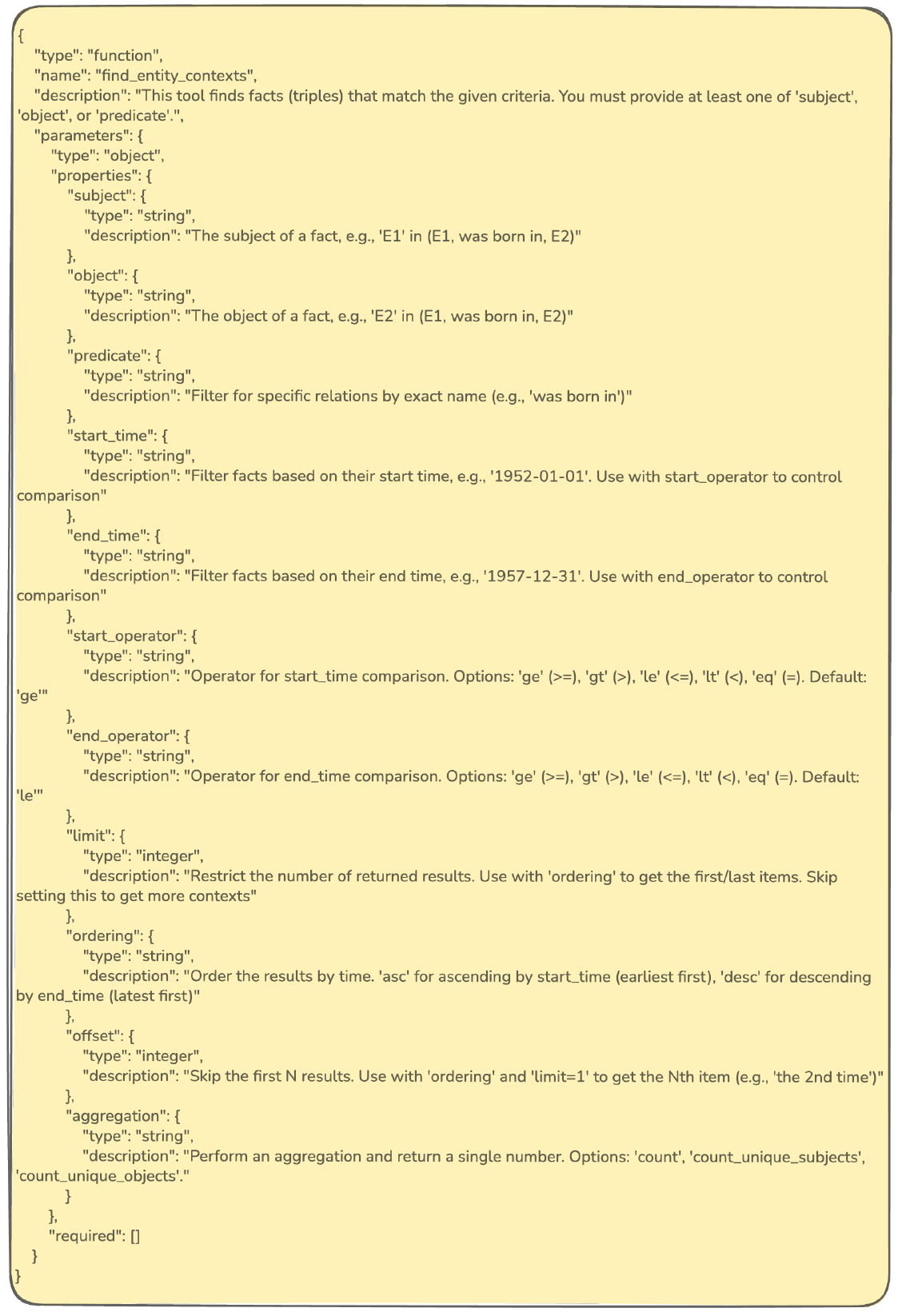}
    \caption{The tool descriptions for \textit{find\_entity\_contexts}.}
    \label{fig:tool description 3}
\end{figure}

This section shows details of \ours.
The prompts used for gist and fact extraction are illustrated in Figure \ref{fig:gist extraction} and Figure \ref{fig:fact extraction}.
The prompts for tool selection, along with the corresponding tool descriptions, are presented in Figures \ref{fig:tool selection}–\ref{fig:tool description 3}.
The tool selector selects one of the available tools at each step based on their descriptions.

\section{Implementation Details}
\label{sec:appendix-implementation-details}

We replicated using open-source versions of Mem0 \citep{mem0} and Graphiti \citep{zep} rather than proprietary ones, aligning settings as closely as feasible, including the backbone LLM, embedding model, and the scale of contexts.
Since we observed Mem0 frequently autonomously choosing to reject adding input text to memory, we opted for finer granularity: adding memory at the message level to encourage more information to be incorporated into memory.
For Graphiti, we found that frequent LLM and embedding calls incurred excessive time and economic costs, and added memory at the session level. Subsequently, Graphiti indexes information within the session into multiple facts for subsequent retrieval.
Thus, both Mem0 and Graphiti store information at the fact/sentence level rather than the session level.

\section{Analysis}

\subsection{Comparative Analysis: \ours vs. TISER}
\label{subsec:case studies}

We compare \ours with NV-Embed-v2 w/ TISER \citep{tiser} on Complex-TR and show a few examples. NV-Embed-v2 w/ TISER is denoted as TISER in this paragraph.
\ours outperformed TISER typically by handling multi-hop temporal reasoning more comprehensively and recovering the full set of required entities. For example, when asked ``\textit{Where was Nancy L. Ross educated before ASU?}'', the gold answer was ``\textit{BC Cancer Research Centre}'', and ours produced ``\textit{Virginia Tech}'' and ``\textit{BC Cancer Research Centre}'', which the judge accepted as correct, while TISER only returned Virginia Tech. 
Conversely, TISER surpassed \ours, usually by pinpointing the exact target within the specified time window, whereas ours produced over-verbose lists or drifted to the wrong temporal hop. 

For instance, in the query ``\textit{Where was Barack Obama educated after State Elementary School Menteng 01?}'', the gold was ``\textit{Punahou School}''. TISER returned exactly that, while \ours instead listed later universities.
This issue is most likely due to the ambiguity of the terms ``\textit{before}'' and ``\textit{after}'', which may introduce uncertainty in determining whether an inequality should include equality when using the provided tools in \ours, and the resulting tool calls do not always align with the intended meaning of the question.
However, the precise relationship between ``\textit{before}'' and ``\textit{after}'' can often be more reliably inferred from the retrieved context, since it can be directly inferred within the temporal scope of a sequence of events.

\subsection{Time and Space Efficiency}
\label{subsec:efficiency-analysis}


In the indexing phase, Graphiti is two orders of magnitude less efficient than Mem0 and \ours. This may be because its processing of each episode involves multiple rounds of calls to both the generative LLM and the embedding model.
In the inference phase, the single-step runtime of \ours is comparable to Mem0 and Graphiti, while its multi-step runtime grows linearly with the number of steps.

\begin{table}[]
\small
\centering
\caption{The running time and memory usage.}
\label{tab:time-space usage}
\begin{tabular}{lrrr}
\toprule
REMem-I & Index Time (s) & Inference Time / Query (s) & Max Memory Usage (GB) \\ \midrule
LoCoMo & $3,604.2$ & $4.3$ & $2.6$ \\
Complex-TR & $378.7$ & $12.6$ & $1.3$ \\
\bottomrule
\end{tabular}
\end{table}

We further report the runtime and memory usage of REMem in Table \ref{tab:time-space usage}, where it reports the memory consumption of REMem itself. If an LLM or embedding model service is deployed locally, its main memory and GPU memory usage are counted separately.
Experiments were conducted on a server with dual AMD EPYC 7643 48-Core CPUs (96 hardware threads), 4× NVIDIA A100-SXM4-80GB GPUs, and 1 TB of system memory.
The inference time for each query is affected by many factors, including the number of workers and the response time of the LLM service.
The reported time excludes any LLM or embedding caching and does not use multithreading. With caching or multithreading enabled, the actual throughput would be higher.

\subsection{Token Usage}

The token usage of \ours on LoCoMo is shown in Table \ref{tab:token-usage}.
\ours-S exhibits token consumption nearly equivalent to the embedding baseline during the inference phase, while the token consumption of \ours-I increases with the number of iterations.

\begin{table}[tb]
  \centering
  \small
  \caption{Token usage on LoCoMo ($1,986$ queries) using OpenAI o200k\_base encoding. The estimated cost is calculated using the standard GPT-4.1-mini pricing.}
  \begin{tabular}{l|cc|cc|c|c}
    \toprule
    & \multicolumn{2}{c|}{Indexing Phase} & \multicolumn{2}{c|}{Inference Phase} & Total & Est. Cost (USD) \\
    \cmidrule(lr){2-3}\cmidrule(lr){4-5}
    Method & Input & Output & Input & Output & & \\
    \midrule
    NV-Embed-v2  & N/A       & N/A         & \ \ $1.19$M & $0.17$M & \ \ $1.36$M & $\$\ \ 0.75$ \\
    \ours-I      & \ \ $0.90$M & \ \ $0.72$M & $18.10$M & $0.79$M  & $20.51$M & $\$10.02$ \\
    \ours-S      & \ \ $0.90$M & \ \ $0.72$M & \ \ $1.83$M  & $0.18$M & \ \ $3.63$M & $\$\ \ 2.53$ \\
    \bottomrule
  \end{tabular}
  \label{tab:token-usage}
\end{table}

\subsection{Example of the Extraction}

We present a passage from MuSiQue, along with the extraction results from Mem0 and \ours, in Table \ref{tab:extraction example}.
In comparison, Mem0's extraction results exhibit a lack of factual coverage, which may hinder the subsequent comprehension and QA task.
The extraction from \ours is more comprehensive, especially as the last (14th) fact, even inferring that the season close of RFEF (2002-2003) is June 2003.
Mem0 stores memories prefixed with ``user'', assuming the working scenario is always conversational messages, yet this is unnecessary for long passages that describe objective knowledge.


\begin{longtable}{p{\dimexpr\textwidth-2\tabcolsep}}


\caption[]{A MuSiQue passage and its extraction results from Mem0 and \ours.} \\
\toprule
\endfirsthead

\caption[]{A MuSiQue passage and its extraction results from Mem0 and \ours. (Continued)} \\ 
\toprule
\endhead

\midrule
\multicolumn{1}{r}{\textit{Continued on next page}} \\
\endfoot

\bottomrule
\label{tab:extraction example} \\

\endlastfoot

\textbf{Passage: Lionel Messi} \\
\midrule
After a year at Barcelona's youth academy, La Masia, Messi was finally enrolled in the Royal Spanish Football Federation (RFEF) in February 2002. Now playing in all competitions, he befriended his teammates, among whom were Cesc Fàbregas and Gerard Piqué. After completing his growth hormone treatment aged 14, Messi became an integral part of the `Baby Dream Team'', Barcelona's greatest-ever youth side. During his first full season (2002--03), he was top scorer with 36 goals in 30 games for the Cadetes A, who won an unprecedented treble of the league and both the Spanish and Catalan cups. The Copa Catalunya final, a 4--1 victory over Espanyol, became known in club lore as the partido de la máscara, the final of the mask. A week after suffering a broken cheekbone during a league match, Messi was allowed to start the game on the condition that he wear a plastic protector; soon hindered by the mask, he took it off and scored two goals in 10 minutes before his substitution. At the close of the season, he received an offer to join Arsenal, his first from a foreign club, but while Fàbregas and Piqué soon left for England, he chose to remain in Barcelona. \\ \midrule
\textbf{Mem0 Extraction} \\
\midrule
User mentioned Lionel Messi's early football career details including his time at Barcelona's youth academy La Masia, enrollment in the Royal Spanish Football Federation in February 2002, friendship with teammates Cesc Fàbregas and Gerard Piqué, completion of growth hormone treatment at age 14, being top scorer with 36 goals in 30 games during the 2002-03 season for Cadetes A, winning a treble of league and Spanish and Catalan cups, playing with a plastic mask after a broken cheekbone and scoring two goals, and receiving an offer from Arsenal but choosing to stay in Barcelona. \\

\midrule

\textbf{\ours Gists} \\ 
\midrule
1. Lionel Messi was enrolled in the Royal Spanish Football Federation (RFEF) in February 2002 after a year at Barcelona's youth academy, La Masia. \\
2. Messi befriended his teammates Cesc Fàbregas and Gerard Piqué while playing in all competitions. \\
3. After completing his growth hormone treatment at age 14, Messi became an integral part of Barcelona's greatest-ever youth side, the 'Baby Dream Team'. \\
4. During his first full season (2002–03), Messi was the top scorer with 36 goals in 30 games for the Cadetes A. \\
5. The Cadetes A won an unprecedented treble of the league and both the Spanish and Catalan cups during the 2002–03 season. \\
6. The Copa Catalunya final, a 4–1 victory over Espanyol, became known as the partido de la máscara, the final of the mask. \\
7. A week after suffering a broken cheekbone during a league match, Messi was allowed to start the Copa Catalunya final on the condition that he wear a plastic protector. \\
8. Messi took off the plastic protector and scored two goals in 10 minutes before his substitution in the Copa Catalunya final. \\
9. At the close of the 2002–03 season, Messi received an offer to join Arsenal, his first from a foreign club. \\
10. Messi chose to remain in Barcelona while Fàbregas and Piqué soon left for England. \\
\midrule

\textbf{\ours Facts} \\ 
\midrule
1. (Lionel Messi, was enrolled in, the Royal Spanish Football Federation (RFEF)) [point in time: 2002-02] \\
2. (Lionel Messi, befriended, Cesc Fàbregas and Gerard Piqué) \\
3. (Lionel Messi, became an integral part of, Barcelona's greatest-ever youth side, the 'Baby Dream Team') [point in time: 2002-03] \\
4. (Lionel Messi, was the top scorer with, 36 goals in 30 games for the Cadetes A) [point in time: 2002-03] \\
5. (Cadetes A, won, an unprecedented treble of the league and both the Spanish and Catalan cups) [point in time: 2002-03] \\
6. (Copa Catalunya final, was a victory over, Espanyol) [point in time: 2002-03] \\
7. (Copa Catalunya final, became known as, the partido de la máscara, the final of the mask) \\
8. (Lionel Messi, suffered, a broken cheekbone during a league match) [point in time: 2002-03] \\
9. (Lionel Messi, was allowed to start, the Copa Catalunya final on the condition that he wear a plastic protector) [point in time: 2002-03] \\
10. (Lionel Messi, took off, the plastic protector) [point in time: 2002-03] \\
11. (Lionel Messi, scored, two goals in 10 minutes before his substitution in the Copa Catalunya final) [point in time: 2002-03] \\
12. (Lionel Messi, received an offer to join, Arsenal, his first from a foreign club) [point in time: 2003-06] \\
13. (Lionel Messi, chose to remain in, Barcelona) \\
14. (Cesc Fàbregas and Gerard Piqué, left for, England) [point in time: 2003-06] \\
\end{longtable}

\subsection{Graph Property}
\label{subsec:graph_property}

For the evaluated four benchmarks, Table~\ref{tab:graph_properties} reports the scale of the constructed memory graphs, including the numbers of phrase and gist nodes, edges, triples, and associated token counts. 
The statistics show that LoCoMo, REALTALK, and Complex-TR induce large, densely connected graphs with many context and synonymy edges.
For Test of Time, the graph is much smaller and contains no gist-level annotations, since its statements involve anonymous entities and are already highly formalized, so we extract only fact-level information.

\begin{table}[htb]
\centering
\small
\caption{The graph properties on evaluated benchmarks: average values for each graph.}
\begin{tabular}{lrrrr}
\toprule
 & LoCoMo & REALTALK & Complex-TR & Test of Time \\
\midrule
\# of phrase nodes                   & $777.5$     & $974.0$     & $1,066.0$   & $16.0$     \\
\# of gist nodes                     & $730.1$     & $889.0$     & $1,095.0$   & --         \\
\# of relation edges                 & $736.2$     & $891.0$     & $1,062.0$   & --         \\
\# of context edges                  & $25,172.4$  & $56,332.0$  & $2,190.0$   & --         \\
\# of synonymy edges                 & $1,082.2$   & $606.0$     & $748.0$     & --         \\
\# of triples                        & $763.9$     & $917.0$     & $1,095.0$   & $275.4$    \\
\midrule
\# of input tokens                   & $15,965.8$  & $19,250.0$  & $26,158.0$  & $4,109.7$  \\
\# of phrase-node tokens & $5,105.8$   & $8,350.0$   & $5,536.0$   & $32.1$     \\
\# of gist-node tokens   & $21,743.8$  & $30,106.0$  & $26,514.0$  & --         \\
\midrule
Node degree of phrase nodes          & $34.0$      & $59.7$      & $4.1$       & $9.7$      \\
Node degree of gist nodes            & $34.4$      & $63.4$      & $2.0$       & --         \\
\bottomrule
\end{tabular}
\label{tab:graph_properties}
\end{table}

\end{document}